%% file: acl_latex.tex
\pdfoutput=1

\documentclass[11pt]{article}

\usepackage[preprint]{acl}

\usepackage{times}
\usepackage{latexsym}
\usepackage{multirow}
\usepackage{amsmath}
\usepackage{booktabs}
\usepackage{graphicx} 
\usepackage{adjustbox} 
\usepackage{rotating}
\usepackage[T1]{fontenc}

\usepackage[utf8]{inputenc}
\usepackage{comment}

\usepackage{microtype}

\usepackage{inconsolata}

\usepackage{graphicx}

\usepackage{changepage,threeparttable} 

\input{macros}

\title{Steering off Course: Reliability Challenges in Steering Language Models}

\author{Patrick Queiroz Da Silva$^{\clubsuit}$ \quad Hari Sethuraman$^{\heartsuit}$ \\ \textbf{Dheeraj Rajagopal}$^{\spadesuit}$ \quad \textbf{Hannaneh Hajishirzi}$^{\heartsuit \diamondsuit}$ \quad \textbf{Sachin Kumar}$^{\clubsuit}$ \\
$^\clubsuit$The Ohio State University, Columbus OH \quad 
$^\heartsuit$University of Washington, Seattle WA \\
$^\spadesuit$Fastino AI, Palo Alto CA \quad
$^\diamondsuit$Allen Institute for AI, Seattle WA \\
\texttt{\small dasilva.30@osu.edu}
}

\begin{document}
\maketitle
\begin{abstract}
Steering methods for language models (LMs) have gained traction as lightweight alternatives to fine-tuning, enabling targeted modifications to model activations. However, prior studies primarily report results on a few models, leaving critical gaps in understanding the robustness of these methods. In this work, we systematically examine three prominent steering methods---DoLa, function vectors, and task vectors. In contrast to the original studies, which evaluated a handful of models, we test up to 36 models belonging to 14 families with sizes ranging from 1.5B to 70B parameters. Our experiments reveal substantial variability in the effectiveness of the steering approaches, with a large number of models showing no improvement and at times degradation in steering performance. Our analysis demonstrate fundamental flaws in the assumptions underlying these methods, challenging their reliability as scalable steering solutions\footnote{full code, data, and results can be found on our \href{https://github.com/patqdasilva/steering-off-course}{GitHub}}.

\end{abstract}

\section{Introduction}
Building on a growing array of interpretability tools \citep{zhao2023explainabilitylargelanguagemodels, ferrando2024primerinnerworkingstransformerbased, rai2024practicalreviewmechanisticinterpretability}, \textit{steering} methods for language models have gained popularity as a way to modify model behavior with specific objectives at inference time \cite{vig2020causal, meng2022locating, li2023inferencetime}. They have been applied to steer models toward desirable outputs, such as improved factuality \citep{chaudhary2024evaluating,zhao2024steering} or away from undesirable traits \citep{o2024steering,farrell2024applying}.
These techniques are appealing because they require little data compared to fine-tuning and do not require changes to model parameters \citep{subramani-etal-2022-extracting,turner2023activation,rimsky-etal-2024-steering}. 
However, they are still poorly understood and face significant challenges that hinder their practical applicability.

\begin{figure}[t]
  \includegraphics[width=\columnwidth]{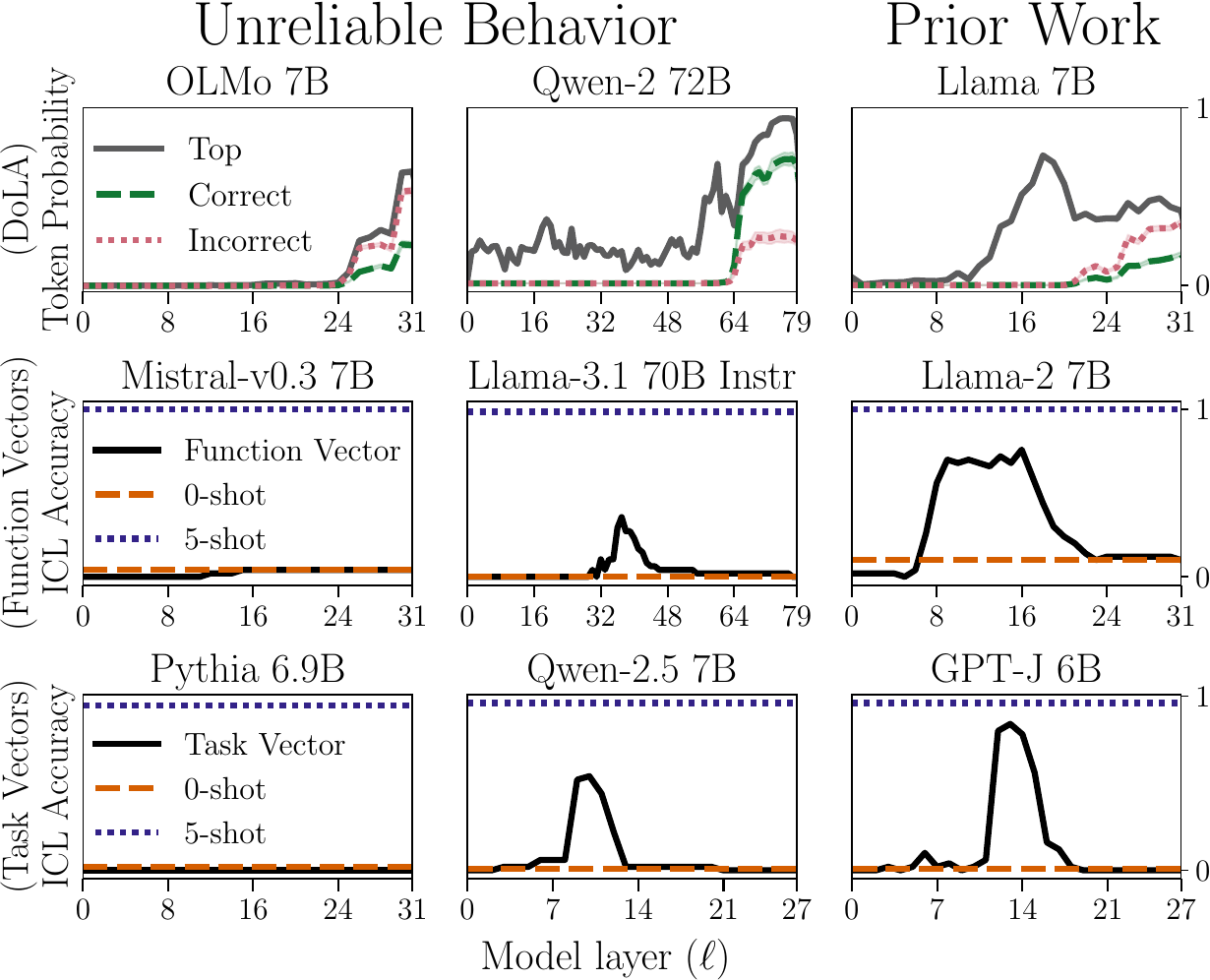}
  \caption{We study the generalization of various LM steering methods to previously unstudied models and find high variance in steering performance. (Top plot) Building on logit lens \Sref{sec:VD}, DoLa contrasts token probabilities across layers to discover factual answers. We show that models do not have similar patterns, limiting the effectiveness of this method. (Middle and bottom plots) Function vectors and task vectors are two methods for steering based on activation patching \Sref{sec:activation}. We show that activation patching results in highly variable performance across many model families and sizes.}
  \label{fig:fig1}
\end{figure}

The LM steering literature has accumulated a generalization blind spot. Most prior work reports results on only a select number of LMs \cite{geva-etal-2022-transformer, gurnee2023finding, zou2023representationengineeringtopdownapproach, chuang2024dola, templeton2024scaling, huben2024sparse}, with growing evidence of unintended side effects. Recent work has shown that steering methods can be brittle \citep{tan2025analyzinggeneralizationreliabilitysteering} and harm general LM capabilities \citep{durmus2024steering}.

Motivated by these studies, we adopt a critical perspective of two popular interpretability tools and the steering methods they inspired. Specifically, logit lens~\citep{logitlens} and activating patching \citep{vig2020causal}. We focus on three steering methods based on these tools: DoLa \citep{chuang2024dola} built on logit lens (\Sref{sec:VD}), and function vectors \citep{todd2024function} and task vectors \citep{hendel2023incontext} built using ideas from activation patching (\Sref{sec:activation}). The original studies evaluated a small number of LMs; our primary goal is to quantify the generalization of these methods across different model families and sizes.
Our experiments show a large variance in performance across 36 decoder-only transformer-based LMs from 14 model families with sizes ranging from 1.5B to 70B parameters. Several model families, even after significant hyperparameter tuning, show no improvement or even decline in relevant steering metrics. Our analyses reveal that these steering methods rely on flawed assumptions about the generalization of internal transformer mechanisms. We postulate many more hypotheses for such variance and provide recommendations for the future of evaluating steering methods for LMs in \Sref{sec:discussion}.

\section{Related Work}

\paragraph{Brittleness of Steering Methods}

Recent work has highlighted the brittleness of different steering methods providing motivation for this work. For example, sparse autoencoders (SAEs) aim to learn sparse, interpretable features from internal activations of LMs \citep{elhage2022toy}. The features can be upweighted or downweighted to steer the models \citep{huben2024sparse}. \citet{durmus2024steering} show that targeted steering often induces unpredictable behaviors harming the general capabilities of the model. Related to this work, \citet{wu2025axbenchsteeringllmssimple} show that SAEs are not effective in steering compared to simple baselines like prompting. Additionally, \citet{tan2025analyzinggeneralizationreliabilitysteering} study steering vectors based on contrastive activation addition (CAA) \citep{rimsky-etal-2024-steering}. They focus on out-of-distribution generalization and show that some concepts in LMs are unsteerable. Most related to our work is \citet{brumley2024comparingbottomuptopdownsteering}, who report inconsistencies in function and in-context vector steering capabilities across different kinds of tasks. In contrast to our work, they experiment with only two mid-sized models. We test the generalization of steering methods across many model families and sizes with extensive hyperparameter search, and subsequently make a case against the underlying assumptions that steering vectors are built upon.

\paragraph{Brittleness of Model Editing}
Similar to steering methods, model editing methods do not require training to modify model behavior. Instead of updating model activations at inference time, model editing aims to locate and update model weights responsible for a specific behavior, such as storing facts \cite{meng2022locating}.
Similar to our exploration of steering, prior work has shown that the brittleness and side effects of model editing can limit its practical usability. \citet{li2024unveiling} demonstrate two of these unintended consequences, knowledge \textit{conflicts} and \textit{distortions}, in popular locate-then-edit knowledge editing tools \citep{mitchell2022fastmodeleditingscale, meng2022locating, meng2023massediting}. \citet{yao-etal-2023-editing,huang2024knowledgeeditingreallycorrect} explore the generalization of knowledge editing and find that edits can struggle to maintain consistency across contexts, often hurting generalization.

\section{Logit Lens} \label{sec:VD}
Introduced in \citet{logitlens},  the logit lens provides insights into how LMs refine their prediction across layers (or sublayers). This approach has been applied to interpret activations at various stages \citep{geva2020transformer}, as well as to discover circuits \citep{lieberum2023does,wang2022interpretability}.
Specifically, the output of any model layer, $h_\ell$, can be projected onto the vocabulary space to obtain logits by multiplying it by the unembedding matrix ($W_U$), $\mathrm{LogitLens}(h_\ell) = h_\ell W_U$, optionally followed by softmax to convert it to a probability distribution over the vocabulary (also known as probits), $q_\ell (\cdot) = \mathrm{softmax} (\mathrm{LogitLens}(h_\ell))$.
The tokens with the highest logit values or probabilities can be used to infer what information has been encoded in the hidden layer to the extent that the information is linearly decodable in the space defined by the model's unembedding matrix. In addition to understanding model dynamics, logit lens has also been used to steer model behavior based on insights from said dynamics~\citep{bhalla2024towards}. For example, to reduce hallucinations \citep{Gema2024DeCoReDB,Chen2024LowerLM,jiang2024devils}. In this section, we analyze DoLA \citep{chuang2024dola}, one of the most influential works on logit-lens-based steering that focuses on improving factuality. 

Central to this work is the hypothesis that neurons that store factual knowledge are distributed among the later layers of the model \cite{dai-etal-2022-knowledge}. During inference time, while these neurons contribute to an increase in the probability of factually correct outputs, this increase may not be sufficient to ensure that the probability of the correct output is the greatest among all possible outputs. Based on this hypothesis, rather than using the absolute probability of the following token, \citet{chuang2024dola} compute the \emph{relative change} in probability at the final layer compared to an earlier or ``premature'' layer. The layer $\ell$ farthest from the final layer $L$, in terms of Jensen-Shannon Divergence (JSD) between $q_L$ and $q_\ell$, is chosen as the premature layer.  

More formally, for a model with $L$ layers, the premature layer at inference step $t$ is chosen as $P=\arg \max_{\ell \in \mathcal{B}} \mathrm{JSD}(q_L(\cdot \mid x_{<t}) \parallel q_\ell(\cdot \mid x_{<t}))$. Here, $\mathcal{B}$ refers to the set of all ``candidate'' layers (except the final). The candidate set, henceforth referred to as buckets, is a hyperparameter. Using $P$, the output probability is updated as 
\begin{align}
 \hat{p}(x_t|x_{<t}) = \mathrm{softmax} (\mathcal{F}(q_L,q_P))
 \label{eq:postsoftmax}
\end{align}
where
\begin{align}
  \mathcal{F}(q_L,q_P)_{x_t}  &= 
    \begin{cases}
      \log \frac{q_L(x_t)}{q_P(x_t)} & \text{if } x_t \in \mathcal{V}_{head}\\
      -\infty & \text{otherwise}
    \end{cases}
    \label{eq:alpha}
\end{align}

where $\mathcal{V}_{head}$ is the set of tokens $x_t$ for which $q_P(x_t)\ge \alpha \max_w q_L(w)$. Here $\alpha$ is another hyperparameter. 
As $\alpha$ increases, a larger portion of tokens with low probabilities are set to have a probability of zero. We refer to the readers to \citet{chuang2024dola} for addit details.

\subsection{Experimental Setup} 
\paragraph{Tasks, datasets, and models}
Following the original paper, we evaluate DoLa on two multiple choice text completion tasks: TruthfulQA \cite{tfqa}\footnote{While there exists a generative version of TruthfulQA, it requires GPT3-based evaluation which is no longer offered by OpenAI APIs}  and FACTOR \cite{factor}. TFQA measures the truthfulness of models in answering questions. It consists of questions spanning various domains to test models' tendencies to generate false but human-plausible answers. FACTOR measures the factuality of a model. We experiment with the ``News'' subset. The input consists of a prefix followed by either a correct completion or one of multiple incorrect completions, where the probability of each completion is computed.

We evaluate with a comprehensive set of 10 models that span 7 model families and two scales. We experiment with Llama 1 at 7B \cite{touvron2023llamaopenefficientfoundation}, Llama 2 at 7B and 70B \cite{touvron2023llama2openfoundation},  Llama 3 at 8B and 70B \cite{grattafiori2024llama3herdmodels}, Pythia  at 6.9B \cite{biderman2023pythiasuiteanalyzinglarge}, Mistral v0.1 at 7B \cite{jiang2023mistral7b}, OLMo at 7B \cite{groeneveld2024olmoacceleratingsciencelanguage}, and Qwen 2 at 7B and 72B \cite{yang2024qwen2technicalreport}.

\paragraph{Hyperparameters}
DoLa uses two primary hyperparameters that may affect its final performance. The first is the choice of $\mathcal{B}$, the set of candidate layers from which the premature layer is chosen, also referred to as the bucket. The optimal bucket is typically chosen using a validation set. In our experiments, we report results over 4 buckets defined by a range of layers\footnote{There are in principle an exponential number of potential buckets, we focus on ranges of layers following prior work.}. For the small models, our buckets are the bottom 50\% of the layers (0-50\%), middle 50\% of the layers (25-75\%), top 50\% of the layers (50-100\%), and all layers (except the final). For the large models, our buckets are the first (0-25\%), second (25-50\%), third (50-75\%), and last (75-100\%) quartiles of layers. For each bucket, following the original paper, we only consider even-numbered layers (for efficiency reasons).

The other hyperparameter is $\alpha$, which determines $\mathcal{V}_\text{head}$ (see \autoref{eq:alpha}). $\alpha$ adjusts the threshold in token probability of the mature layer, below which probabilities are set to zero. To avoid a computationally expensive grid search over all buckets and $\alpha$ values for each model, we first run a search over the buckets, with $\alpha=0.1$, the value used in \citet{chuang2024dola}. Once we determine the best bucket for each model, we search over  $\alpha \in \{0, 0.1, 0.25, 0.5, 0.75, 0.9 \}$ values.

\paragraph{Evaluation}
For both tasks, we compute 6-shot performance for the baseline approach (without any steering) and DoLA. For TruthfulQA, we compute three metrics following prior work: MC1, MC2, and MC3. The task specifies different answer choices to the model depending on the metric to be computed. For MC1, only one of the options is the correct answer. MC1 measures the accuracy of a model's greedy prediction using the probabilities of different answer choices. For MC2 and MC3, more than one answer may be correct. MC2 measures the normalized score assigned to the set of correct answers. MC3 defines the ranked-choice prediction accuracy---whether the set of correct answers is assigned a higher likelihood than the wrong ones. Regardless of the input, a 6-shot prompt, where each example consists of a question followed by one correct answer, is added as a prefix. For FACTOR, we compute accuracy (same as MC1).

We also note that \citet{chuang2024dola} report these metrics by treating $\mathcal{F}$ as log-probabilities (which they are not) without computing a softmax and using $\hat{p}$ (using \autoref{eq:postsoftmax}). Since all these metrics are computed by aggregating $\hat{p}$ over multiple tokens, directly aggregating logits as output by $\mathcal{F}$ can lead to a length bias where, depending on the signs of logits, a longer output may unfairly be rewarded or punished (more details are provided in \Aref{asec:dola_results}). We report all our results using $\hat{p}$.

\begin{table}
  \centering
  \resizebox{\columnwidth}{!}{
  \begin{tabular}{lccccccc}
    \toprule
    \multirow{2}{2cm}{\textbf{Model}} & \multicolumn{2}{c}{\textbf{MC1}} & \multicolumn{2}{c}{\textbf{MC2}} & \multicolumn{2}{c}{\textbf{MC3}}\\
    \cline{2-7}
    & \textbf{Base} & \textbf{DoLa} & \textbf{Base} & \textbf{DoLa} &\textbf{Base} & \textbf{DoLa} \\
    \midrule
    LLama 7B* & 0.26 & 0.32 & 0.41 & 0.64 & 0.19 & 0.32 \\ 
    Llama 7B &  0.26 & 0.32 & 0.41 & 0.52 & 0.19 & 0.28 \\
    LLama 2 7B & 0.29 & 0.29 & 0.43 & 0.44 & 0.21 & 0.21\\
    Llama 3 8B & 0.32 & 0.32 & 0.49 & 0.49 & 0.24 & 0.24\\
    Pythia 6.9B & 0.23 & 0.26 & 0.37 & 0.52 & 0.18 & 0.23\\
    Mistralv0.1 7B & 0.32 & 0.32 & 0.48 & 0.48 & 0.24 & 0.24\\
    Qwen 2 7B & 0.39 & 0.37 & 0.58 & 0.51 & 0.30 & 0.30\\
    OLMo 7B& 0.25 & 0.25 & 0.40 & 0.40 & 0.19 & 0.19\\
    Llama 2 70B & 0.37 & 0.34 & 0.54 & 0.54 & 0.27 & 0.27\\
    Llama 3 70B & 0.35 & 0.35 & 0.56 & 0.56 & 0.28 & 0.28\\
    Qwen 2 72B & 0.44 & 0.39 & 0.63 & 0.46 & 0.33 & 0.30\\
    \bottomrule
  \end{tabular}
  }
  \caption{Performance of DoLA for TruthfulQA with the best hyperparameter combination (chosen based on the highest MC1 value). The results for the full search can be found in \Aref{asec:dola_results}. * refers to the DoLa results reported by \citet{chuang2024dola} using $\mathcal{F}$ without applying a softmax.}
  
  \label{tab:tfqa}
\end{table}

\subsection{Results and Analysis}
\label{subsec:dola_results}

We summarize our main results in \autoref{tab:tfqa} and \autoref{tab:factor_results} for TFQA and FACTOR respectively. Each row corresponds to the best result chosen based on MC1 for TFQA and accuracy for FACTOR. We provide complete results for all buckets and $\alpha$ values in \Aref{asec:dola_results}.  

\citet{chuang2024dola} report results only on Llama 1 family of models. We note that, after correctly computing the metrics (i.e. using $\hat{p}$ instead of $\mathcal{F}$), the performance improvement offered by DoLa over the baseline approach is \emph{much} lower,  although still substantial (more details in \autoref{tab:tfqa_all} in \Aref{asec:dola_results}). However, for every other model we evaluate, with the exception of Pythia, the improvements afforded by DoLa in most metrics is not significant, especially compared to the gains of Llama 1. Additionally, in models such as Qwen 2 7B and Llama 2 70B, we see a slight deterioration in some metrics compared to the baseline. The hyperparameters (buckets and $\alpha$) have little effect on this trend (see Tables~\ref{tab:tfqa_bucket_small}, \ref{tab:tfqa_bucket_large}. \ref{tab:tfqa_alpha_search}, \ref{tab:factor_bucket_search_small}, \ref{tab:factor_bucket_search_large}, and \ref{tab:factor_alpha_search} in \Aref{asec:dola_results}). Our results are consistent with several follow-up works that have attempted to apply this approach on top of select fine-tuned models and achieved little or no improvements \cite{tian2023fine,pan2025lisa}.

\begin{table*}[h]
    \centering
    \setlength\tabcolsep{2pt}
    \resizebox{\textwidth}{!}{
    \begin{tabular}{lrrrrrrrrrrr}
        \toprule
        & \multicolumn{1}{c}{\textbf{LLama 7B*}} & \multicolumn{1}{c}{\textbf{Llama 7B}} & \multicolumn{1}{c}{\textbf{Llama 2 7B}} & \multicolumn{1}{c}{\textbf{Llama 3 8B}} & 
        \multicolumn{1}{c}{\textbf{Pythia 6.9B}} & \multicolumn{1}{c}{\textbf{Mistralv0.1}} &
        \multicolumn{1}{c}{\textbf{Qwen 2 7B}} & \multicolumn{1}{c}{\textbf{OLMo}} &
        \multicolumn{1}{c}{\textbf{Llama 2 70B}} & \multicolumn{1}{c}{\textbf{Llama 3 70B}} & \multicolumn{1}{c}{\textbf{Qwen 2 72B}} \\
        \midrule
        \textbf{Base} & 58 & 58 & 72 & 76 & 51 & 76 & 69 & 67 & 83 & 85 & 82 \\
        \textbf{DoLa} & 62 & 63 & 73 & 76 & 48 & 76 & 68 & 67 & 82 & 86 & 82 \\
        \bottomrule
    \end{tabular}
    }
    \caption{Results (\% accuracy) measuring the performance of DoLA for FACTOR with the best hyperparameter combination (results for the full search can be found in \Aref{asec:dola_results}). * refers to the DoLa results reported by \citet{chuang2024dola} using $\mathcal{F}$ without applying a softmax.}
    \label{tab:factor_results}
\end{table*}

\begin{figure}[t]
  \includegraphics[width=\columnwidth]{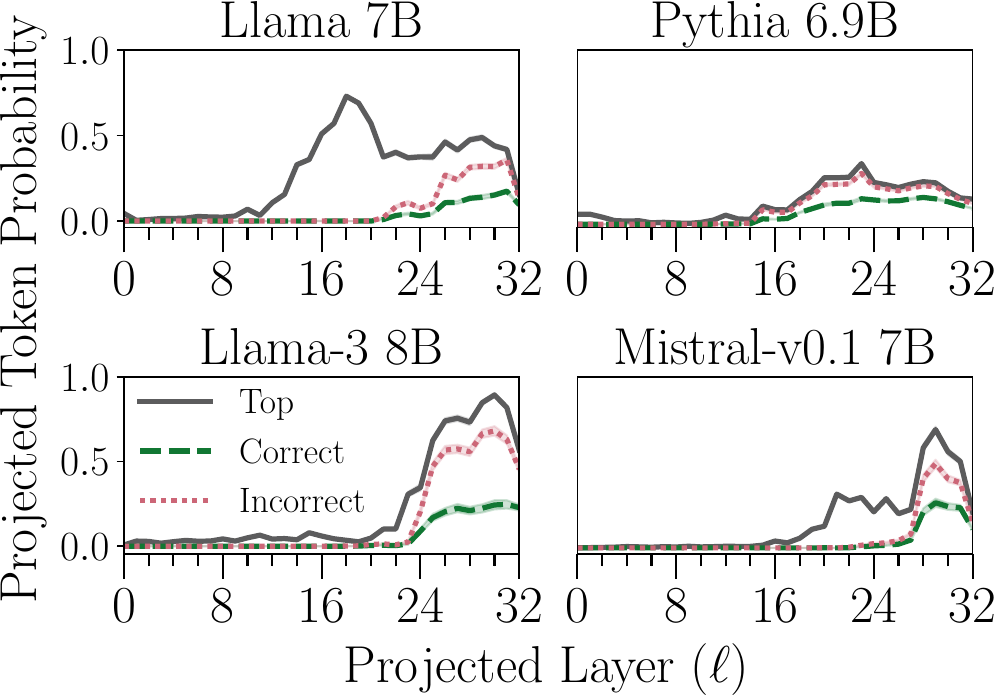}
  \caption{Projected token probabilities from hidden states at each layer of 4 selected LMs on the TruthfulQA dataset (the remaining results can be found in \Aref{asec:logit_lens_analysis}). The correct and incorrect token probabilities begin spiking at the same layer, which suggests that a contrast with early layers would be relatively uninformative.}
  \label{fig:logit_lens}
\end{figure}

We hypothesize that the failure of this approach stems from the flawed assumption that ``factual knowledge of an LM evolves gradually across layers'' \citep{meng2022locating,geva-etal-2023-dissecting}, 
and that contrasting the final with earlier layers offers a meaningful signal. \citet{wiegreffe2025answer} show that in a multiple choice setting, there exist model-specific layers that start the ``promotion'' of the answer tokens. Inspired by their methodology, we measure the probability of 3 token types across each layer using TruthfulQA: the correct and incorrect final answers, and the token with the highest logit value. \autoref{fig:logit_lens} aligns with the results from \citet{wiegreffe2025answer}, showing that the correct and incorrect tokens have low probabilities before spiking at the same layer. This suggests that a contrast with early layers is relatively uninformative, especially when comparing probabilities of correct and incorrect tokens. This result contrasts with DoLa's motivation, which states that the promotion of early tokens should be discounted to encourage more of what the model prefers in later layers.

\section{Activation Patching}
\label{sec:activation}

Activation patching is the technique of replacing or updating internal activations of a neural network with another vector to modify a specific behavior of the model \citep{vig2020causal,geiger2020neural,geiger2021causal,meng2022locating, chan2022causal, wang2023interpretability}. It is typically used to interpret models by isolating task-specific circuits and observing relevant changes in model outputs~\citep{hanna2023how, heimersheimjaniak2023, conmy2023towards}, but it has also been applied to model steering. Previous work has explored several ways to create \textit{steering} vectors~\citep{nanda-etal-2023-emergent, zou2023representationengineeringtopdownapproach, tigges2023linearrepresentationssentimentlarge, turner2024steering}. At inference time, model activations at different layers are added to or replaced by these vectors to change the properties of the generated text.

In this section, we examine two related activation patching techniques which have been used to study in-context learning (ICL) in transformers: \textbf{function vectors} (FVs)~\citep{todd2024function} and \textbf{task vectors} (TVs)~\citep{hendel2023incontext}. For a task $t$, test example $\tilde{x_i}$, set of exemplars $S^K_t=\{(x_1, y_1), \ldots, (x_K, y_K)\}$ demonstrating the task, and prompt $p^{t, K}_i = [S^K_{t}, \tilde{x_i}] \in P_t$, both techniques aim to compress the demonstrations into a steering vector $v_t(S)$. When patched into an LM, this vector, in theory, should reproduce the performance of ICL in a zero-shot setup. 

To create FVs, \citet{todd2024function} rely on the localization hypothesis \citep{olsson2022context}, suggesting
the that ``presence of a handful of attention heads in an LM can mediate many ICL tasks.'' The FV is computed in two steps. First, the average activation of a head over all prompts from a task dataset is defined as $\bar{a}^t_{\ell j} = \frac{1}{|P_t|}\sum_{p^t_i \in P_t}a_{\ell j}(p_i^{t, 10})$ where $a_{\ell j}$ is a self-attention head at layer $\ell$. It is used to construct the function vector
\begin{equation}
\label{eq:fv}
v^{FV}_t = \sum_{a_{\ell j} \in \mathcal{A}_n} \bar{a}_{\ell_j}^t
\end{equation}
where $\mathcal{A}_n$ is a subset containing $n$ attention heads ranked by a causal mediation analysis \citep{Pearl_2009} and $n$ is a hyperparameter. At a high level, given ICL prompts with random, uninformative pairs, the causal indirect effect of a head is measured as its ability to recover the correct answer when patched with its mean head activation $\bar{a}^t_{\ell j}$ on the desired task. We report the results for the casual indirect effect of each head in Appendix \ref{ap:fv_form}. For more details on the formulation of FVs, please refer to \citet{todd2024function}.

Task vectors on the other hand do not rely on localization or causal mediation analysis. They directly compress a task into the activation space of a transformer model
\begin{equation}
\label{eq:tv}
v^{TV}_t = h_\ell = f_\ell(p^{t,5}_i)
\end{equation}
where $f_\ell$ is a transformer model producing the hidden state at layer $\ell$. Additional heuristics of TV implementation can be found at \citet{hendel2023incontext}.

Using unseen prompts, the steering vectors $v_t(S)$ can be applied to the hidden states of any layer $\ell$ as 
\begin{align}
\label{eq:act_patch}
h_\ell \leftarrow \alpha h_\ell + \lambda v_t
\end{align}
Here $\lambda$ is typically set to 1 (we also experiment with other values for FVs). $\alpha$ is set to 1 for FVs and 0 for TVs. 

\subsection{Experimental Setup}
\label{sec:ap_setup}
\paragraph{Tasks and datasets}
We source ICL word-pair samples from \citet{hendel2023incontext, todd2024function} and select 11 representative tasks from linguistic, factual knowledge, and translation tasks. This includes generating the antonym of English words, the past tense of a present-tense verb, the capital of a country, and various translations to and from English (eng to [lang], [lang] to eng), using French, Spanish, German, and Italian. 
Full dataset details can be found in \Aref{ap:ap_data}.

\paragraph{Models}
We use 36 models ranging from sizes 1.5B to 70B across the GPT-J \citep{gpt-j}, Pythia (P) \citep{biderman2023pythiasuiteanalyzinglarge}, Llama 1 (L) \citep{touvron2023llamaopenefficientfoundation}, Llama 2 (L2) \citep{touvron2023llama2openfoundation}, Llama 3.1 (L3.1), Llama 3.2 (L3.2) \citep{grattafiori2024llama3herdmodels}, Mistral v0.1 (M1) and v0.3 (M3) \citep{jiang2023mistral7b}, Gemma 2 (G2) \citep{gemmateam2024gemma2improvingopen}, Qwen 2 (Q2) \citep{yang2024qwen2technicalreport}, Qwen2.5 (Q2.5) \citep{qwen2025qwen25technicalreport}, OLMo (O) \citep{groeneveld2024olmoacceleratingsciencelanguage}, OLMo 2 (O2) \citep{olmo20252olmo2furious}, Amber \citep{liu2023llm360fullytransparentopensource}, and Falcon 3 \citep{Falcon3} families.\footnote{Due to changes to caching and tokenization across HuggingFace versions, we do not replicate Gemma 2 and OLMo 2 in task vectors. In function vectors, these models follow similar trends as others. Additional results are provided in \Aref{ap:ap_results}} We use only open-source models, as the experiments require access to model internals.

\paragraph{Hyperparameters}
FVs and TVs both have what could be considered a hyperparameter; the transformer layer $\ell$ at which the steering vector is patched. During inference, we apply the steering vector to only a single layer and search across all layers in the model. Besides $\ell$, TVs do not have any additional hyperparameters.

\citet{todd2024function} only consider one hyperparameter for FVs---$\mathcal{A}_n$, the (number of) top attention heads to use when constructing the FV (as shown in \autoref{eq:fv}). 
While they demonstrate that performance generally saturates in GPT-J after 10 heads, our exploration indicates that other models and tasks may respond differently. We experiment with $n \in \{2, 16, 32, 64, 128, 256, 512, 1024\}$.

We also introduce $\lambda$ (see \autoref{eq:act_patch}), a strength multiplier on the function vector. $v^{FV}_t$ is created with a subset of attention heads and can have a low norm compared to $h_\ell$; therefore, we explore $\lambda \in \{0.5, 1, 2, 4, 8, 16, 32, 64\}$.

In addition to the best performance we obtain with hyperparameter search, we also report FV's performance on \citet{todd2024function}'s default settings. We use $n \in \{2, 16\}$ for small models ($\leq$14B) and $n \in \{2, 64\}$ for large models ($>$14B). We also set a default $\lambda \in \{1\}$, consistent with both task and function vectors.

\paragraph{Evaluation}
For each task, we use a test set of 50 word pairs\footnote{Considering computational budget and extensive hyperparameter search, we use 50 test samples as our hypotheses are only afflicted by large changes in accuracy.} in FV, and the default number of word pairs using the implementation from \citet{hendel2023incontext} for TV. For each of our 11 ICL tasks, we create prompts with K exemplars and use the highest probability token as the prediction following prior work. While the FV is created using $K \in \{10\}$, we record baseline performance using $K \in \{0, 5\}$ to be consistent with the defaults from TV implementation. In our tasks, 10-shot performance tends to marginally improve over 5-shot, thus we underestimate brittleness in FVs (see \autoref{fig:five_ten_shot} in \Aref{ap:ap_evaluation}). To determine steering vector efficacy, we apply activation patching at layer $\ell$ with $K \in \{0\}$ and record the corresponding accuracy.

Due to limited space, we report aggregated results across our three hyperparameters in the main paper (with detailed results in \Aref{ap:ap_results}). We accept $\mathcal{A}_n$ and $\lambda$ to be highly variable across models and thus always take the maximum score across those hyperparameters. Function vectors and task vectors are reported to work over clusters of early/middle layers; therefore, we also report both a peak and average \textbf{performance recovery metric} for layers. In general, we report peak (max) and average (mean) performance recovery as their respective aggregated steering vector performance across layers. We normalize these numbers by the respective model's 5-shot performance. The divisor is intended to remove noise in ICL performance. Exact formulas are defined in \Aref{ap:ap_evaluation}. As a final aggregation, we also report the frequency with which model-task combinations surpass a peak performance recovery at different quantiles of their 5-shot performance.

\begin{table}
  \centering
  \begin{tabular}{lcccc}
    \hline
    \textbf{5-shot Perf.} & \textbf{.50} & \textbf{.75} & \textbf{.90} & \textbf{1.00} \\
    \hline
    FV Default Param   & 47\% & 37\% & 20\% & 12\%       \\
    FV Param Search   & 76\% & 68\% & 52\% & 28\%        \\
    Task Vector         & 69\% & 54\% & 35\% & 16\%         \\ \hline
  \end{tabular}
  \caption{The percent of model-task combinations which surpass a quantile of their respective 5-shot performance using the best combination of hyperparameters: \{$\ell, \lambda, \mathcal{A}_n$\} for FV and \{$\ell$\} for TV.}
  \label{tab:5shot_actpatch}
\end{table}

\begin{figure}[t]
  \includegraphics[width=\columnwidth]{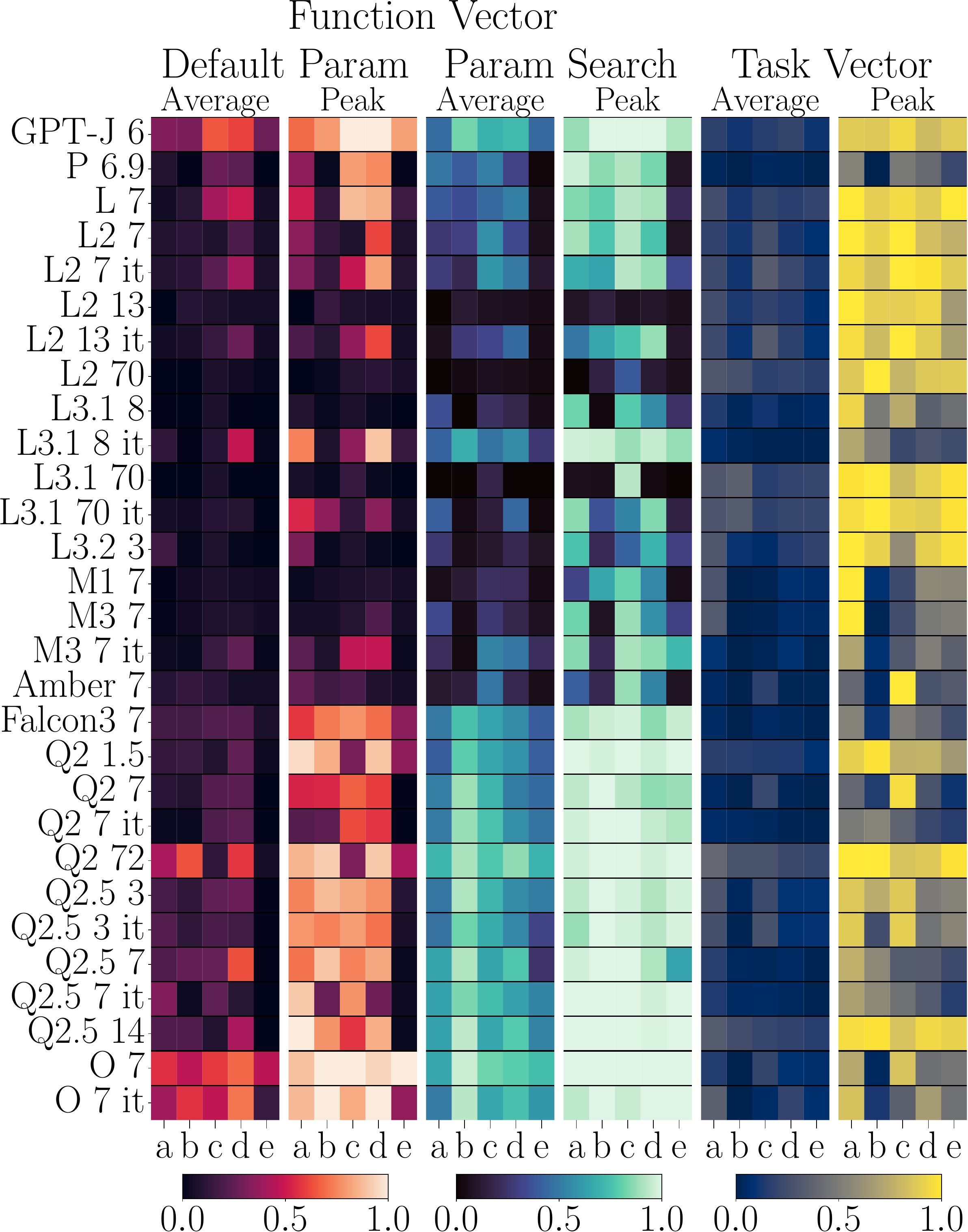}
  \caption{Performance recovery with different activation patching methods across tasks and models. There is large variance in performance across tasks, models, and tools. Tasks: a) Antonym, b) Present-Past, c) Country-Capital, d) [lang] to eng, and e) eng to [lang].}
  \label{fig:6plots}
\end{figure}

\subsection{Results and Analysis}
\label{sec:ap_results}
\begin{figure}[t]
  \includegraphics[width=\columnwidth]{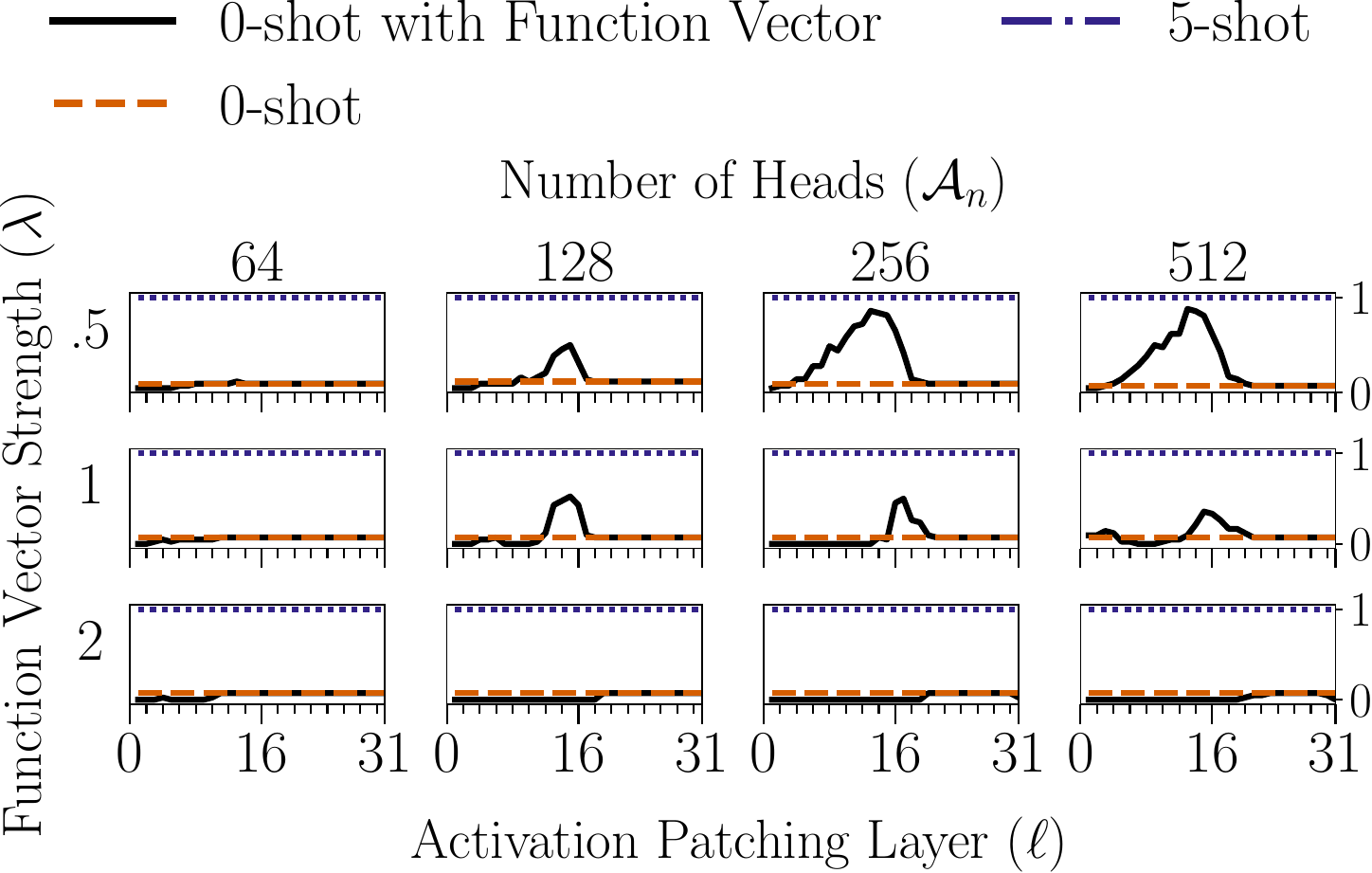}
  \caption{A subspace in the hyperparameter search across ($\ell, \lambda, \mathcal{A}_n$) for Mistral-v0.3 7B on the Country-Capital task. The FV does not become effective until 128 (10\%) heads, and it continues to grow in performance until 512 (50\%) heads. This provides some evidence against localization for this model and task, as many heads are required before performance emerges.}
  \label{fig:fv_param_search}
\end{figure}

\begin{figure}[t]
  \includegraphics[width=\columnwidth]{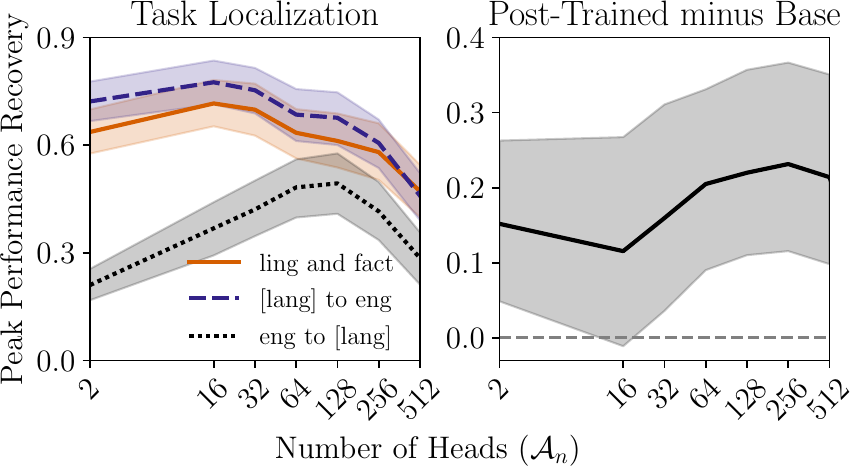}
  \caption{(Left) Some tasks work well with few heads, while translating from English requires more heads to be effective. (Right) Post-trained models outperform their base counterparts on average, especially when more heads are used to construct the function vector. However, there is significant variance when averaging across models and tasks.}
  \label{fig:head_task_instr}
\end{figure}

We begin by optimistically quantifying the brittleness of function vectors and task vectors, summarized in \autoref{tab:5shot_actpatch} and \autoref{fig:6plots}. We analyze the interaction between hyperparameters and activation patching, summarized in Figures~\ref{fig:fv_param_search} and \ref{fig:head_task_instr}.

\paragraph{Neither FVs nor TVs are generalizable}
As shown in \autoref{tab:5shot_actpatch}, even allowing for noise, FVs with default parameters recover 5-shot performance in only 20\% of model-task combinations, and 52\% with parameter search across $\mathcal{A}_n$ and $\lambda$; TVs perform poorly as well at 35\%. Even with the best-performing tool, FVs with full hyperparameter search, only 76\% of model-task combinations reach 50\% of 5-shot performance.

Additionally, aggregate results from Figure \ref{fig:6plots} demonstrate the low performance and high variability across models and tasks. Function vectors have nearly no pattern in their efficacy, besides eng to [lang] having poor steerability in many models. We find TV to be more stable across larger scales. All of the seven larger ($\geq$13B) models we tested worked decently with TV.

We further compare base and post-trained models. Using FV, post-trained models perform better on average, especially when the base model performs poorly as demonstrated in Figures \ref{fig:6plots} and \ref{fig:head_task_instr}. However, with TVs, we observe a decline in performance with post-trained models (see \fref{fig:tv_instr} in \Aref{ap:best_param}). The uneven impact of post-training on function and task vectors provides further evidence for the fragility of steering based on activation patching.

\paragraph{Impact of function vector strength ($\lambda$) and layer ($\ell$)}
We address the assumption that the FV should have a norm strong enough to impact the residual stream with $\lambda$. We generally note that models prefer some range of $\lambda$ (some lower and some higher), as shown in \autoref{fig:best_lambda} in \Aref{ap:best_param}, although there is still moderate variability across tasks. Some models work with the default $\lambda \in \{1\}$, but varying this hyperparameter helped to recover performance in many models.

Both TVs and FVs assume that there is at least one layer $\ell$ in which the model will respond to the steering vector. However, not only are some models ineffective at all layers, the range of best patching layers varies widely across models and tasks.
\autoref{fig:best_layer} in \Aref{ap:best_param} provides a comprehensive summary of the preferred patching layer for every model and task. Additionally, the high peak but low average recovery for task vectors visible in \autoref{fig:6plots} suggests a strong dependence on the choice of $\ell$ for task vectors.

\paragraph{Localization hypothesis does not always hold for FVs} 
FVs rely on the assumption that information required for ICL is stored and activated within a small subset of heads, which we test with $\mathcal{A}_n$. We find that on some tasks, certain models require many heads in their FV before recovering performance, with one example displayed in \autoref{fig:fv_param_search}.

The translation to English, linguistic, and factual knowledge tasks show some evidence for localization. They are most effective when $n$ is relatively small, as shown in \autoref{fig:head_task_instr}. However, when translating from English to another language, the more localized $n \in \{2, 16, 32\}$ have lower performance, and performance improves with $n \in \{64, 128\}$. 

Whether a model is post-trained also has an interaction with $\mathcal{A}_n$. Figure \ref{fig:head_task_instr} highlights that post-trained models outperform base models at higher $n$. This result could suggest that the instruction-tuning process produces models with more distributed access to information.

We summarize the variable levels of steerability with differing number of attention heads $n$ across all models and tasks in \autoref{fig:best_heads} in \Aref{ap:best_param}.

\section{Discussion}
\label{sec:discussion}
Our findings show a clear breakdown in steering tools across models and confusing differences in dynamics between seemingly similar models. In this section, we enumerate several potential hypotheses that may explain these differences. While we conduct some experiments to verify them, our limited resources preclude us from making a conclusive case for any of them.

\paragraph{Model pretraining}
Recent models have adopted a two-stage pretraining process \citep{blakeney2024does}, also referred to as mid-training \citep{olmo20252olmo2furious}, involving a second round of training on higher quality raw corpus to increase the context window \citep{gao2024train} or improve later instruction following abilities for tasks \citep{olmo20242}. We hypothesize that the second stage, which is performed with a much lower learning rate, may result in late-spiking logit lens dynamics, and hence poor results for techniques such as DoLa, as shown in \autoref{fig:logit_lens}.

To test this hypothesis, we use OLMo, which at the time of running these experiments was the only model with publicly accessible weights at different checkpoints. However, we find no significant differences in dynamics for checkpoints at the end of stage one and two. In fact, we find OLMo starts showing late spiking dynamics very early in the pretraining stage at 100B tokens, as shown in \autoref{fig:logit_lens_olmo} in \Aref{asec:logit_lens_analysis}. However, we cannot conclusively refute this hypothesis with experiments with only one model.

\paragraph{Model architecture and optimization}
We investigated several architectural differences among LMs: number of layers and attention heads, number of key-value heads, vocabulary size, hidden dimension sizes, attention types, activation functions, and context length. Model differences are summarized in \autoref{tab:model_arch} and \autoref{tab:model_dto}. Among these model differences, we find no discernible pattern of brittleness caused by any such changes. One exception is Gemma 2 9B which has been trained via knowledge distillation and shows vastly different dynamics than other models \citep{gemmateam2024gemma2improvingopen}. For example, in logit lens, Gemma2 9B has a top token probability of nearly 1.0 in early layers, demonstrated in \autoref{fig:logit_lens_other} in \Aref{asec:logit_lens_analysis}. We acknowledge that all models we looked at are trained with different training datasets and optimization pipelines. With more resources, future work may conduct controlled experiments to ablate these differences.

\paragraph{Training data} Finally, we speculate that the number of tokens, data quality, and style may play a role in model differences. For example, recent models disobey the Chinchilla scaling laws \citep{10.5555/3600270.3602446} and saturate the models with many tokens. It is conceivable that the amount of training tokens supplied to the model plays a role in how interpretable their mechanics are.

\section{Conclusion}
We study the generalization of three popular steering methods: DoLa, function vectors, and task vectors. We find that all three techniques are extremely brittle across models and tasks. Our analysis reveals that the underlying assumptions upon which these methods are based are flawed. This work adds to the growing evidence of robustness challenges that surgical methods aiming to modifying language model behavior face. A large amount of interpretability research from which such steering methods are derived continues to be shared via informal channels such as blog posts with seemingly exciting results but limited formal evaluation. As such, we implore future research in this direction to adopt a rigorous evaluation setup considering a wide array of models and tasks to test the reliability of steering approaches, as models with minor differences may have vastly different manifestations of key behavior.

\section*{Limitations}

\paragraph{Non-exhaustive hyperparameter search}
While we performed an extensive hyperparameter search for all steering methods, we did not cover all possible combinations due to our limited computational budget. For example, for DoLa, there are a prohibitively large number of bucket combinations from which the premature layer can be selected, while we experiment with a smaller set. We did, however, find very similar results across all bucket combinations providing confidence that other combinations are unlikely to improve the results.

\paragraph{Non-exhaustive model and task selection}
We experimented with a wide array of popular open-source models, however, at the time of writing this paper, many newer models have been released that we have not tested which may reveal different results. Additionally, for activation patching, we choose a representative sample of tasks from prior work. It is possible that there are other tasks are more steerable than we report, and some worse. However, the main message of the paper remains the same: there exists high variability in the performance of the steering methods we tested.

\paragraph{Variance explanation} We observe that the behaviors which steering methods build upon manifest in many forms and formulate multiple potential hypotheses in \Sref{sec:discussion}. Wherever possible, we conduct experiments to test the hypotheses (such as with OLMo checkpoints and Gemma 2 public training details). However, due to our limited resources, we are ultimately unable to run controlled experiments to conclusively verify or refuse any of the proposed hypotheses.

\section*{Acknowledgments}
The authors would like to thank Sarah Wiegreffe, Noah A. Smith, Vidhisha Balachandran, and Yulia Tsvetkov for helpful discussions and feedback on the early draft of this paper.

\bibliography{acl_latex}

\appendix

\section{Additional Details and Results for Logit Lens}
In this section, we provide additional experimental details, results, and analysis to complement \Sref{sec:VD}.

\subsection{Results}
\label{asec:dola_results}

\paragraph{Length bias in original DoLa results}
In \autoref{tab:tfqa} in \Sref{subsec:dola_results}, we discuss that \citet{chuang2024dola} uses $\mathcal{F}$ (\autoref{eq:alpha}) to compute the reported metrics treating them as log-probabilities (when, in fact, they are similar to logits). Using $\mathcal{F}$ gives the illusion that DoLA improves on the baseline. We find that this happens because $\mathcal{F}$ results in logit values that are all $>0$. Hence, an option which is longer in length (in terms of number of tokens) gets a higher score as the logits add up. This length bias leads to several instances in TruthfulQA where the right answer is the longest being predicted correctly.

We show that without determining any premature layers, we can artificially increase or decrease TruthfulQA performance by adding or subtracting a constant value from the logits at the final layer and using the modified logits for the prediction without applying a softmax. To introduce this bias, we first subtract the smallest logit value from all logits to make them all positive. As shown in \autoref{tab:tfqa_all}, adding a positive constant leads to higher MC1, MC2, and MC3 values and vice versa. We determined that setting this constant to $20$ yielded the highest improvement.

\paragraph{Results for all hyperparameters}
Our preliminary analysis revealed that a lower alpha almost always led to better performance. Hence, with $\alpha=0.1$ (also the value used by \citet{chuang2024dola}), we report results for all buckets we experimented with in Tables~\ref{tab:tfqa_bucket_small}, \ref{tab:tfqa_bucket_large}, \ref{tab:factor_bucket_search_small} and \ref{tab:factor_bucket_search_large}. For the best bucket found for each model, we show results over different values of $\alpha$ in Tables~\ref{tab:tfqa_alpha_search} and \ref{tab:factor_alpha_search}. We report the best results (according to MC1) in the main paper (\autoref{tab:tfqa} and \autoref{tab:factor_results}).

\begin{table*}[h]
    \centering
    \begin{tabular}{lcccc}
    \hline
        \textbf{Model} & \textbf{0-50\%} & \textbf{25-75\%} & \textbf{50-100\%} & \textbf{0-100\%} \\
    \hline
    \multicolumn{5}{c}{\textbf{MC1}} \\
    \hline
        Llama 7B & 0.2375 & 0.2436 & \textbf{0.2546} & 0.2436 \\
        Llama 2 7B & 0.2558 & 0.2497 & 0.2534 & \textbf{0.2595} \\
        Llama 3 8B & 0.2913 & 0.2876 & 0.2864 & \textbf{0.2913} \\
        Pythia 6.9B & 0.2277 & 0.2179 & 0.2093 & \textbf{0.2301} \\
        Mistralv0.1 7B & 0.2987 & 0.2974 & 0.2974 & \textbf{0.2987} \\
        Qwen 2 7B & 0.3550 & \textbf{0.3733} & 0.3623 & 0.3623 \\
        OLMo 7B & 0.2436 & 0.2399 & 0.2411 & \textbf{0.2436} \\
        
        \hline
         \multicolumn{5}{c}{\textbf{MC2}} \\
        \hline
        Llama 7B & 0.4061 & 0.4166 & 0.4414 & 0.4234 \\
        Llama 2 7B & 0.4340 & 0.4350 & 0.4408 & 0.4362 \\
        Llama 3 8B & 0.5129 & 0.5128 & 0.5113 & 0.5129 \\
        Pythia 6.9B & 0.4213 & 0.4213 & 0.5097 & 0.4417 \\
        Mistralv0.1 7B & 0.4841 & 0.4842 & 0.4842 & 0.4841 \\
        Qwen 2 7B & 0.4935 & 0.5073 & 0.4756 & 0.4756 \\
        OLMo 7B & 0.4176 & 0.4205 & 0.4193 & 0.4176 \\

        \hline
         \multicolumn{5}{c}{\textbf{MC3}} \\
        \hline
        Llama 7B & 0.1739 & 0.1797 & 0.1934 & 0.1824 \\
        Llama 2 7B & 0.1935 & 0.1914 & 0.1929 & 0.1954 \\
        Llama 3 8B & 0.2253 & 0.2236 & 0.2231 & 0.2253 \\
        Pythia 6.9B & 0.1782 & 0.1733 & 0.1848 & 0.1796 \\
        Mistralv0.1 7B & 0.2249 & 0.2245 & 0.2245 & 0.2249 \\
        Qwen 2 7B & 0.2840 & 0.2988 & 0.2963 & 0.2963 \\
        OLMo 7B & 0.1779 & 0.1781 & 0.1779 & 0.1779 \\
        \hline
    \end{tabular}
    \caption{Bucket search for TFQA using $\alpha = 0.1$ (small models). Bolded are the best scores for MC1 in each model, which is later used for the alpha search. Continue reading in \Sref{subsec:dola_results}.}
    \label{tab:tfqa_bucket_small}
\end{table*}

\begin{table*}[h]
    \centering
    \begin{tabular}{lcccc}
    \hline
        \textbf{Model} & \textbf{0-25\%} & \textbf{25-50\%} & \textbf{50-75\%} & \textbf{75-100\%} \\
    \hline
    \multicolumn{5}{c}{\textbf{MC1}} \\
    \hline
        Llama 2 70B & 0.3317 & 0.3305 & 0.3341 & \textbf{0.3403} \\
        Llama 3 70B & \textbf{0.2840} & 0.2803 & 0.2712 & 0.2778 \\
        Qwen 2 72B & 0.3599 & 0.3550 & \textbf{0.38807} & 0.2791 \\
        
        \hline
         \multicolumn{5}{c}{\textbf{MC2}} \\
        \hline
        Llama 2 70B & 0.5503 & 0.5559 & 0.5742 & 0.5368 \\
        Llama 3 70B & 0.5106 & 0.5072 & 0.5064 & 0.5081 \\
        Qwen 2 72B & 0.6000 & 0.5959 & 0.5968 & 0.5112 \\
        \hline
         \multicolumn{5}{c}{\textbf{MC3}} \\
        \hline
        Llama 2 70B & 0.2319 & 0.2339 & 0.2474 & 0.2711 \\
        Llama 3 70B & 0.2210 & 0.2201 & 0.2143 & 0.2246 \\
        Qwen 2 72B & 0.2659 & 0.2731 & 0.2895 & 0.2740 \\
        \hline
    \end{tabular}
    \caption{Bucket search for TFQA using $\alpha = 0.1$ (large models). Bolded are the best scores for MC1 in each model, which is later used for the alpha search. Continue reading in \Sref{subsec:dola_results}.}
    \label{tab:tfqa_bucket_large}
\end{table*}

\begin{table*}[h]
    \centering
    \begin{tabular}{lccccccc}
        \hline
        Model & 0.0 & 0.1 & 0.25 & 0.5 & 0.75 & 0.9 & Bucket used \\
        \hline
        \multicolumn{8}{c}{\textbf{MC1}} \\
        \hline
        Llama 7B & \textbf{0.3182} & 0.2546 & 0.2558 & 0.2301 & 0.2387 & 0.2472 & 50-100\% \\
        Llama 2 7B & \textbf{0.2889} & 0.2595 & 0.2570 & 0.2595 & 0.2472 & 0.2399 & 0-100\%\\
        Llama 3 8B & \textbf{0.3170} & 0.2913 & 0.2717 & 0.2448 & 0.2399 & 0.2387 & 0-100\% \\
        Pythia 6.9B & \textbf{0.2607} & 0.2301 & 0.2007 & 0.2167 & 0.2154 & 0.2142 & 0-100\% \\
        Mistralv0.1 7B & \textbf{0.3158} & 0.2987 & 0.2889 & 0.2852 & 0.2766 & 0.2729 & 0-100\% \\
        Qwen 2 7B & 0.3133 & \textbf{0.3733} & 0.3599 & 0.3207 & 0.3170 & 0.3060 & 25-75\% \\
        OLMo 7B & \textbf{0.2509} & 0.2436 & 0.2472 & 0.2399 & 0.2448 & 0.2411 & 0-100\% \\
        Llama 2 70B & 0.2962 & \textbf{0.3403} & 0.3280 & 0.3366 & 0.3378 & 0.3354 & 75-100\% \\
        Llama 3 70B & \textbf{0.3537} & 0.2840 & 0.2656 & 0.2338 & 0.2203 & 0.2203 & 0-25\% \\
        Qwen 2 72B & 0.3329 & 0.3807 & 0.3819 & \textbf{0.3917} & 0.3782 & 0.3672 & 50-75\% \\
        \hline
        \multicolumn{8}{c}{\textbf{MC2}} \\
        \hline
        Llama 7B & 0.5202 & 0.4414 & 0.4968 & 0.5066 & 0.5528 & 0.5562  & 50-100\% \\
        Llama 2 7B & 0.4386 & 0.4362 & 0.4429 & 0.4758 & 0.5067 & 0.5126 & 0-100\% \\
        Llama 3 8B & 0.4884 & 0.5129 & 0.5142 & 0.4998 & 0.4901 & 0.4882 & 0-100\% \\
        Pythia 6.9B & 0.5253 & 0.4417 & 0.4682 & 0.4981 & 0.5043 & 0.5133 & 0-100\% \\
        Mistralv0.1 7B & 0.4817 & 0.4841 & 0.5021 & 0.5171 & 0.5317 & 0.5329 & 0-100\% \\
        Qwen 2 7B & 0.5420 & 0.5073 & 0.4744 & 0.4803 & 0.4902 & 0.4826 & 25-75\% \\
        OLMo 7B & 0.3952 & 0.4176 & 0.4453 & 0.4764 & 0.5093 & 0.5074 & 0-100\% \\
        Llama 2 70B & 0.5255 & 0.5368 & 0.5293 & 0.5467 & 0.5420 & 0.5426 & 75-100\% \\
        Llama 3 70B & 0.5580 & 0.5106 & 0.4753 & 0.4859 & 0.4855 & 0.4744 & 0-25\% \\
        Qwen 2 72B & 0.5771 & 0.5968 & 0.5099 & 0.4599 & 0.4486 & 0.4480 & 50-75\% \\
        \hline
        \multicolumn{8}{c}{\textbf{MC3}} \\
        \hline
        Llama 7B & 0.2803 & 0.1934 & 0.1900 & 0.1745 & 0.1798 & 0.1847 & 50-100\% \\
        Llama 2 7B & 0.2070 & 0.1954 & 0.1852 & 0.1844 & 0.1760 & 0.1693 & 0-100\% \\
        Llama 3 8B & 0.2389 & 0.2253 & 0.2140 & 0.1881 & 0.1815 & 0.1786 & 0-100\% \\
        Pythia 6.9B & 0.2448 & 0.1796 & 0.1671 & 0.1717 & 0.1763 & 0.1745 & 0-100\% \\
        Mistralv0.1 7B & 0.2394 & 0.2249 & 0.2107 & 0.2100 & 0.2043 & 0.2033 & 0-100\% \\
        Qwen 2 7B & 0.2911 & 0.2988 & 0.2850 & 0.2588 & 0.2553 & 0.2473 & 25-75\% \\
        OLMo 7B & 0.1874 & 0.1779 & 0.1740 & 0.1729 & 0.1757 & 0.1791 & 0-100\% \\
        Llama 2 70B & 0.2996 & 0.2711 & 0.2631 & 0.2528 & 0.2436 & 0.2447 & 75-100\% \\
        Llama 3 70B & 0.2759 & 0.2210 & 0.2080 & 0.1824 & 0.1644 & 0.1571 & 0-25\% \\
        Qwen 2 72B & 0.3122 & 0.2895 & 0.2870 & 0.2997 & 0.2962 & 0.2862 & 50-75\% \\
        \hline
    \end{tabular}
    \caption{$\alpha$ search for TFQA, using the best bucket from \autoref{tab:tfqa_bucket_small} and \autoref{tab:tfqa_bucket_large}. Bolded are the highest scores MC1 scores across alpha for each model. Continue reading in \Sref{subsec:dola_results}.}
    \label{tab:tfqa_alpha_search}
\end{table*}

\begin{table*}
    \centering
    \resizebox{\textwidth}{!}{
    \begin{tabular}{lcccccc}
        \hline
        \textbf{Model} & \textbf{Base}  & \textbf{Best $\alpha$} & \textbf{Best Bucket} & \textbf{DoLa} & \textbf{DoLa w/o softmax} & \textbf{Base w/ length bias (+20)} \\
        \hline
        \multicolumn{7}{c}{\textbf{MC1}} \\
        \hline
            Llama 7B & 0.2570 & 0.0 & 50-100\% & 0.3182 & 0.3158 & 0.3182 \\
            Llama 2 7B & 0.2852 & 0.0 & 0-100\% &  0.2889 & 0.3207 & 0.3354 \\
            Llama 3 8B & 0.3195 & 0.0 & 0-100\% & 0.3170 & 0.3317 & 0.3427 \\
            Pythia 6.9B & 0.2252 &  0.0 & 0-100\% & 0.2607 & 0.2962 & 0.3023 \\
            Mistralv0.1 7B & 0.3158 & 0.0 & 0-100\% & 0.3158 & 0.3354 & 0.3623 \\
            Qwen 2 7B & 0.3550 &  0.1 & 25-75\% & 0.3733 & 0.3439 & 0.3831 \\
            OLMo 7B & 0.2509 &  0.0 & 0-100\% & 0.2509 & 0.3121 & 0.3439 \\
            Llama 2 70B & 0.3500 & 0.1 & 75-100\% & 0.3403 & 0.2974 & 0.3341 \\
            Llama 3 70B & 0.3684 & 0.0 & 0-25\% & 0.3537 & 0.3305 & 0.3513 \\
            Qwen 2 72B & 0.4418 & 0.5 & 50-75\% & 0.3917 & 0.4015 & 0.4039 \\
        \hline
        \multicolumn{7}{c}{\textbf{MC2}} \\
        \hline
            Llama 7B & 0.4055 & 0.0 & 50-100\% & 0.5202 & 0.6176 & 0.6151 \\
            Llama 2 7B & 0.4340 & 0.0 & 0-100\% & 0.4386 & 0.6239 & 0.6182 \\
            Llama 3 8B & 0.4884 & 0.0 & 0-100\%  & 0.4884 & 0.6335 & 0.6489 \\
            Pythia 6.9B & 0.3717 &  0.0 & 0-100\% & 0.5253 & 0.5968 & 0.5829 \\
            Mistralv0.1 7B & 0.4814 &  0.0 & 0-100\% & 0.4817 & 0.6282 & 0.6433 \\
            Qwen 2 7B & 0.4935 & 0.1 & 25-75\% & 0.5073 & 0.6474 & 0.6617 \\
            OLMo 7B & 0.3957 & 0.0 & 0-100\% & 0.3952 & 0.6117 & 0.5187 \\
            Llama 2 70B & 0.5240 & 0.1 & 75-100\% & 0.5368 & 0.4749 & 0.6182 \\
            Llama 3 70B & 0.5817 &  0.0 & 0-25\% & 0.5580 & 0.6480 & 0.6476 \\
            Qwen 2 72B & 0.6256 &  0.5 & 50-75\% & 0.4599 & 0.5704 & 0.6804 \\
        \hline
        \multicolumn{7}{c}{\textbf{MC3}} \\
        \hline
            Llama 7B & 0.1924 & 0.0 & 50-100\% & 0.2803 & 0.3010 & 0.2995 \\
            Llama 2 7B & 0.2076 & 0.0 & 0-100\% & 0.2070 & 0.3048 & 0.3064 \\
            Llama 3 8B & 0.2392 & 0.0 & 0-100\% & 0.2389 & 0.3210 & 0.3276 \\
            Pythia 6.9B & 0.1787 & 0.0 & 0-100\% & 0.2448 & 0.3054 & 0.2787 \\
            Mistralv0.1 7B & 0.2389 & 0.0 & 0-100\% & 0.2394 & 0.3177 & 0.3225 \\
            Qwen 2 7B & 0.2840 & 0.1 & 25-75\% & 0.2988 & 0.3297 & 0.3443 \\
            OLMo 7B & 0.1878 & 0.0 & 0-100\% & 0.1874 & 0.3027 & 0.2588 \\
            Llama 2 70B & 0.2526 & 0.1 & 75-100\% & 0.2711 & 0.2202 & 0.3063 \\
            Llama 3 70B & 0.2917 & 0.0 & 0-25\% & 0.2759 & 0.3249 & 0.3286 \\
            Qwen 2 72B & 0.3272 & 0.5 & 50-75\% & 0.2997 & 0.2974 & 0.3527 \\
        \hline
    \end{tabular}
    }
    \caption{TFQA best results for DoLa compared with baseline, DoLa w/o softmax, and baseline w/ length bias. Continue reading in \Sref{subsec:dola_results}.}
    \label{tab:tfqa_all}
\end{table*}

\begin{table*}[h]
    \centering
    \begin{tabular}{lcccc}
        \hline
        Model & 0-50\% & 25-75\% & 50-100\% & 0-100\% \\
        \hline
        Llama 7B & \textbf{0.6253} & 0.5927 & 0.5367 & 0.5784 \\
        Llama 2 7B & 0.7281 & 0.7220 & 0.7169 & \textbf{0.7281} \\
        Llama 3 8B & \textbf{0.7546} & 0.7454 & 0.7536 & 0.7536 \\
        Pythia 6.9B & \textbf{0.4643} & 0.3996 & 0.3697 & 0.4054 \\
        Mistralv0.1 7B & 0.7607 & 0.7597 & 0.7597 & \textbf{0.7607} \\
        Qwen 2 7B & \textbf{0.6782} & 0.6049 & 0.5031 & 0.5061 \\
        OLMo 7B & 0.6640 & \textbf{0.6650} & 0.6629 & 0.6640 \\
        \hline
    \end{tabular}
    \caption{FACTOR bucket search (small models), using $\alpha=0.1$. Continue reading in \Sref{subsec:dola_results}.}
    \label{tab:factor_bucket_search_small}
\end{table*}

\begin{table*}[h]
    \centering
    \begin{tabular}{lcccc}
        \hline
        \textbf{Model} & \textbf{0-25\%} & \textbf{25-50\%} & \textbf{50-75\%} & \textbf{75-100\%} \\
        \hline
        Llama 2 70B & \textbf{0.8214} & 0.8195 & 0.7635 & 0.7124 \\
        Llama 3 70B & 0.8552 & 0.8562 & \textbf{0.8571} & 0.8514 \\
        Qwen 2 72B & \textbf{0.8079} & 0.8012 & 0.7056 & 0.6187 \\
        \hline
    \end{tabular}
    \caption{FACTOR bucket search (large models), using $\alpha=0.1$. Continue reading in \Sref{subsec:dola_results}.}
    \label{tab:factor_bucket_search_large}
\end{table*}

\begin{table*}[ht]
    \centering
    \resizebox{\textwidth}{!}{
    \begin{tabular}{lcccccccc}
        \hline
        \textbf{Model} & \textbf{Baseline} & \textbf{0.0} & \textbf{0.1} & \textbf{0.25} & \textbf{0.5} & \textbf{0.75} & \textbf{0.9}  & \textbf{Bucket used}\\
        \hline
        Llama 7B & 0.5845 & 0.5876 & \textbf{0.6253} & 0.6059 & 0.5784 & 0.5815 & 0.5550 & 0-50\% \\
        Llama 2 7B & 0.7220 & 0.7200 & \textbf{0.7281} & 0.7138 & 0.6772 & 0.6568 & 0.6568 & 0-100\% \\
        Llama 3 8B & 0.7556 & \textbf{0.7587} & 0.7546 & 0.7230 & 0.7037 & 0.7006 & 0.6701 & 0-50\%\\
        Pythia 6.9B & 0.5125 & 0.2828 & 0.4643 & \textbf{0.4797} & 0.4788 & 0.4537 & 0.4614 & 0-50\%\\
        Mistralv0.1 7B & 0.7576 & \textbf{0.7617} & 0.7607 & 0.7301 & 0.6792 & 0.6833 & 0.6701 & 0-100\%\\
        Qwen 2 7B & 0.6925 & \textbf{0.6843} & 0.6782 & 0.6680 & 0.6619 & 0.6191 & 0.6171 & 0-50\%\\
        OLMo 7B & 0.6660 & 0.6619 & \textbf{0.6650} & 0.6415 & 0.6436 & 0.6242 & 0.6375 & 25-75\% \\
        Llama 2 70B & 0.8320 & 0.8176 & \textbf{0.8214} & 0.7983 & 0.7905 & 0.7664 & 0.7317 & 0-25\% \\
        Llama 3 70B & 0.8475 & 0.8494 & \textbf{0.8571} & 0.8127 & 0.7751 & 0.7847 & 0.7558 & 50-75\%\\
        Qwen 2 72B & 0.8224 & \textbf{0.8243} & 0.8079 & 0.7770 & 0.7375 & 0.7355 & 0.7259 & 0-25\% \\
        \hline
    \end{tabular}
    }
    \caption{FACTOR $\alpha$ search using buckets from \autoref{tab:factor_bucket_search_small} and \autoref{tab:factor_bucket_search_large}. Continue reading in \Sref{subsec:dola_results}.}
    \label{tab:factor_alpha_search}
\end{table*}

\subsection{Analysis}
\label{asec:logit_lens_analysis}
\begin{figure}[t]
  \includegraphics[width=\columnwidth]{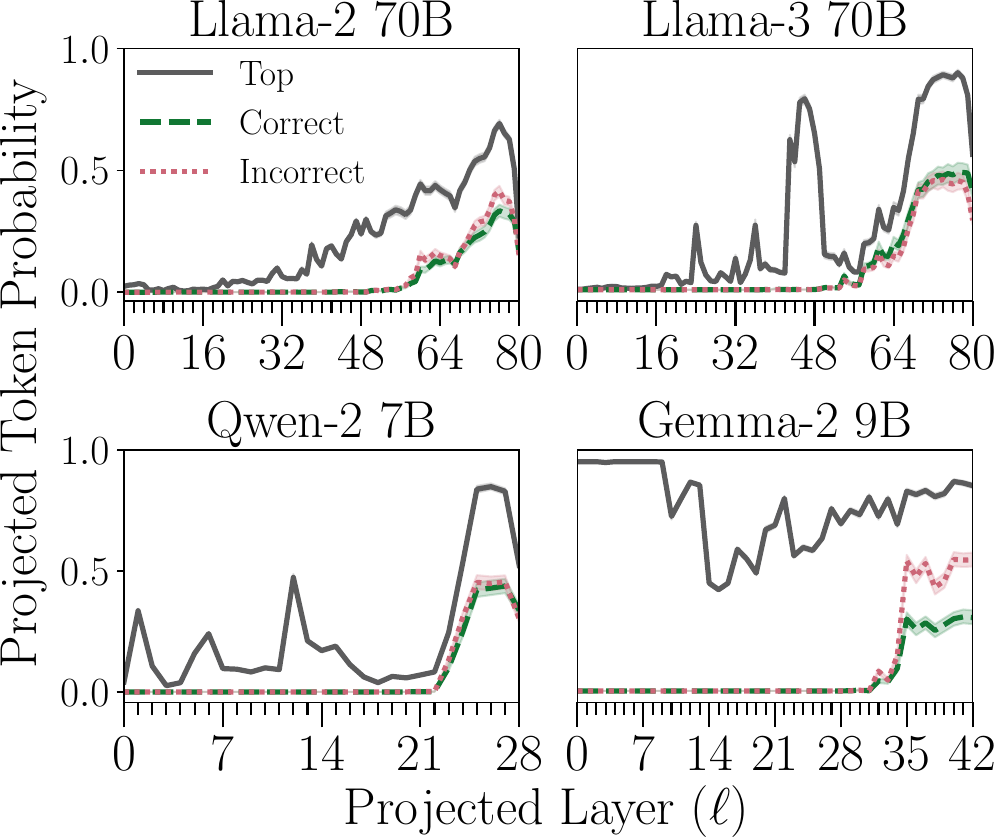}
  \caption{Projected token probabilities from hidden states at each layer of model not reported in the main paper on the TruthfulQA dataset. The correct and incorrect token probabilities begin spiking at the same layer, which suggests that a contrast with early layers would be relatively uninformative. Continue reading in \Sref{subsec:dola_results}.}
  \label{fig:logit_lens_other}
\end{figure}

We provide logit lens analysis plots for all models that are not covered in \autoref{fig:logit_lens} in \autoref{fig:logit_lens_other}.
To further analyze the disparity in the DoLA results among different models and to better understand how tokens are promoted across layers, in addition to the logit lens analysis, we present an additional analysis strengthening our hypothesis that early layers in many models do not provide a meaningful signal to compute premature layers.

\begin{figure}[t]
  \includegraphics[width=\columnwidth]{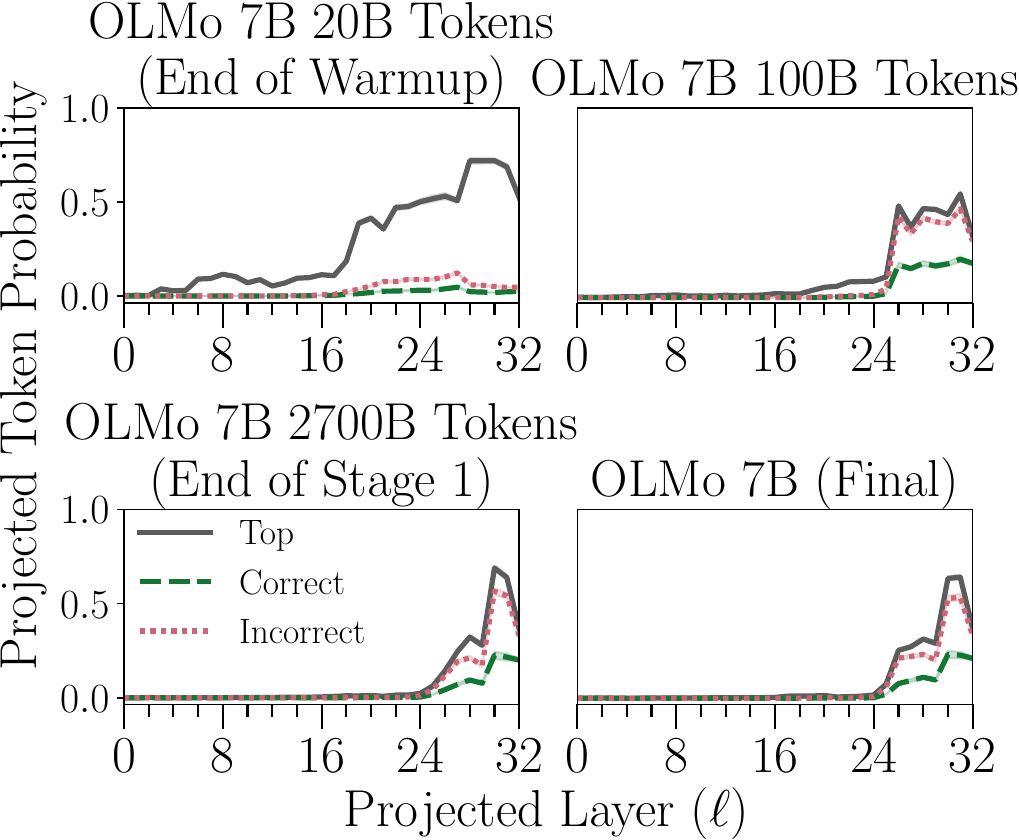}
  \caption{Projected token probabilities from hidden states at each layer of OLMo checkpoints on the TruthfulQA dataset. The correct and incorrect token probabilities begin spiking at the same layer after 100B tokens, suggesting this behavior emerges early in model training, and not with stag 2 of pretraining. Continue reading in \Sref{sec:discussion}.}
  \label{fig:logit_lens_olmo}
\end{figure}

We define a new metric, which we call apathy. 
The primary motive of apathy is to quantify the extent to which transformer components are making changes to the residual stream. The hidden vector output at a layer, $h_\ell$, is a function of the residual stream from the previous layer ($h_{\ell-1}$), and the output from multi-head attention (MHA) and MLP sub-layers. Formally, $h_\ell = h_{\ell-1}  + h_\mathrm{MHA} + h_\mathrm{MLP}$, 
where, $h_\mathrm{MHA} = \mathrm{MHA}(\mathrm{LayerNorm}(h_{\ell-1}))$ and $h_\mathrm{MLP} = \mathrm{MLP}(\mathrm{LayerNorm(h_{\ell-1} + h_\mathrm{MHA})})$.
We define apathy (A) to quantify the resiliency of the residual stream to the additions of MHA and MLP throughout the layers of a transformer:
\begin{align}
    \label{eq:apathy}
    A(r, h) &= (1 +\frac{r \cdot h}{\|r\| \|h\|}) (\|r\|-\|h\|)
\end{align}
$A(h_{\ell-1}, h_\mathrm{MHA})$ corresponds to the contribution of MHA to the residual stream, and $A(h_{\ell-1} + h_\mathrm{MHA}, h_\mathrm{MLP})$ of MLP. Higher apathy means the residual stream is less affected by the outputs of MHA or MLP.

\begin{figure*}[t]
    \centering
  \includegraphics[width=0.7\textwidth]{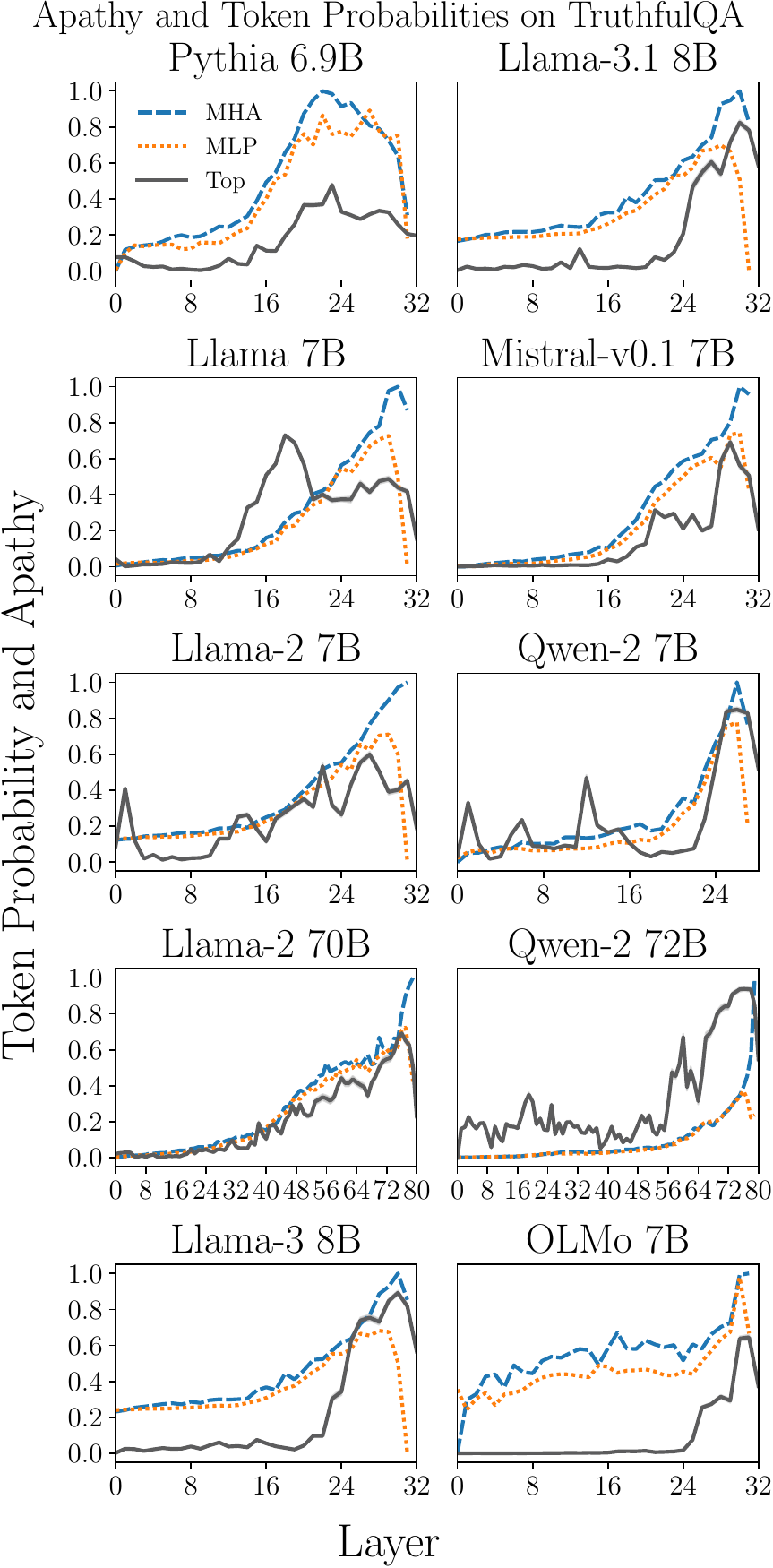}
  \caption{The residual stream starts to ignore MHA and MLP which seems to have some effect on when the top token starts to increase. Each model has its own dynamics, which may help to explain failures using tools which assume that a particular pattern in predicted. Continue reading in \Aref{asec:logit_lens_analysis}.
  }
  \label{fig:apathy}
\end{figure*}

We plot the apathy metric in \fref{fig:apathy}. Some models persist in low apathy for longer, meaning the residual stream is being updated over more layers before it becomes relatively set in norm and direction. The layer at which apathy does begin to increase seems to match where the predicted token promotion occurs, suggesting some interplay between the two. These findings suggest variability in the that way models promote tokens in vocabulary space, making layer contrasting techniques that rely on logit lens (DoLa) unreliable.

Following the discussion in \Sref{sec:discussion}, we show the logit lens analysis for multiple pretraining stages of OLMo in \autoref{fig:logit_lens_olmo}. We demonstrate the emergence of token probability spiking early in training ($\leq100$ billion tokens). We also show the dynamics of Gemma2 9B in \autoref{fig:logit_lens_other}, which has a large token probability with logit lens in its early layers.

\section{Additional Details and Results for Activation Patching}
\label{ap:ap_extra}
In this section, we provide additional experimental details, results, and analysis to complement \Sref{sec:activation}.
\subsection{Additional Details on Tasks and Datasets}
\label{ap:ap_data}
As described in \Sref{sec:ap_setup}, we follow the code implementation which includes full datasets from \citet{hendel2023incontext, todd2024function}. For FVs, we swap source and target languages to get the "[lang] to eng" task, and the equivalent task is available by default in TV.

\subsection{Additional Details on Function Vector Formulation}
\label{ap:fv_form}
\begin{figure*}[t]
  \includegraphics[width=\textwidth]{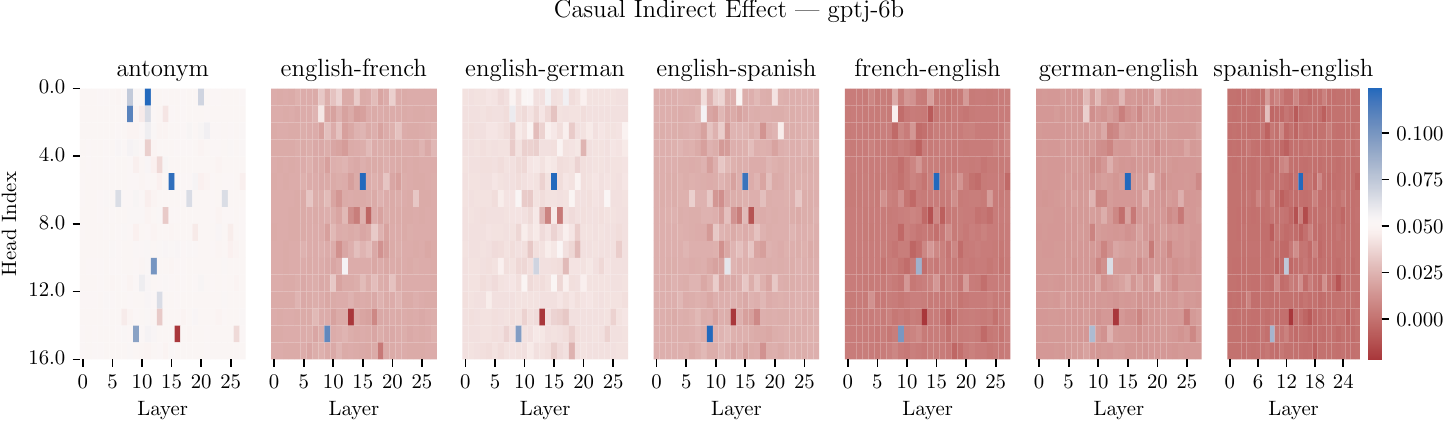}
  \caption{Causal indirect effect of attention heads in GPT-J 6B across all tasks. A few common heads modulate each of its tasks. Continue reading in \Aref{ap:fv_form}.}
  \label{fig:cie_gtpj}
\end{figure*}

\begin{figure*}[t]
  \includegraphics[width=\textwidth]{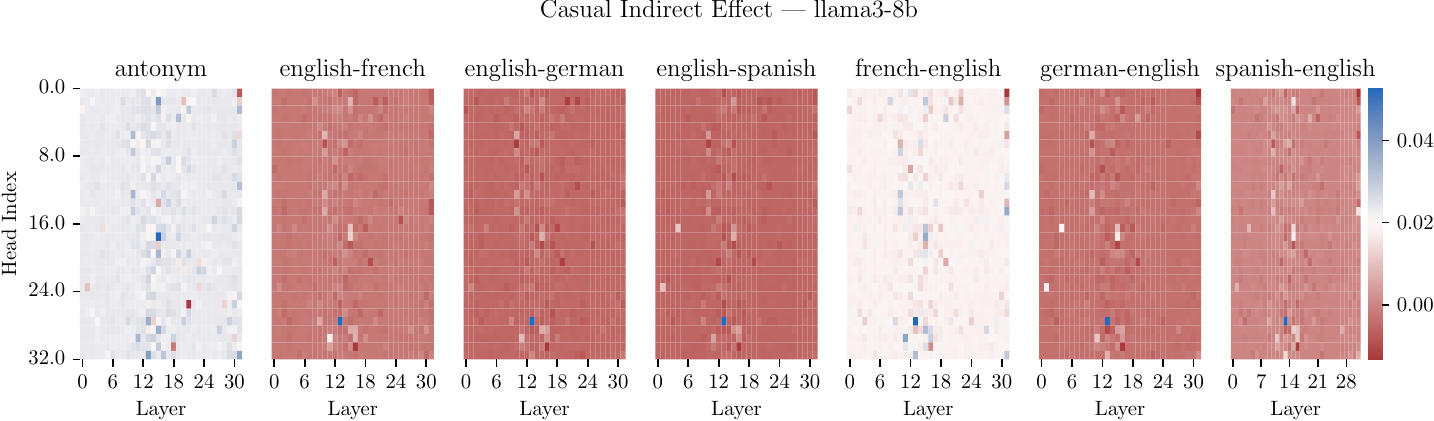}
  \caption{Causal indirect effect of attention heads in Llama-3.1 8B across all tasks. Varied patterns in the heads which modulate each task reflect the variability we observe in our experimental results. Continue reading in \Aref{ap:fv_form}.}
  \label{fig:cie_llama}
\end{figure*}

As shown in \autoref{eq:fv}, finding heads which modulate a task is a key aspect of the casual mediation analysis in function vectors. We share two representative plots for the contribution of each attention head. GPT-J 6B in \autoref{fig:cie_gtpj} shows a few common heads modulate each of its tasks, consistent with \citet{todd2024function}. Llama-3.1 8B in \autoref{fig:cie_llama} shows varied patterns in the heads which modulate each task, reflecting the variability we observe in our experimental results.

To see the causal indirect effect for every model and task, please use the following link: \href{https://github.com/patqdasilva/steering-off-course}{https://github.com/patqdasilva/steering-off-course}

\subsection{Additional Details on Evaluation}
\label{ap:ap_evaluation}
\begin{figure*}[t]
  \includegraphics[width=\textwidth]{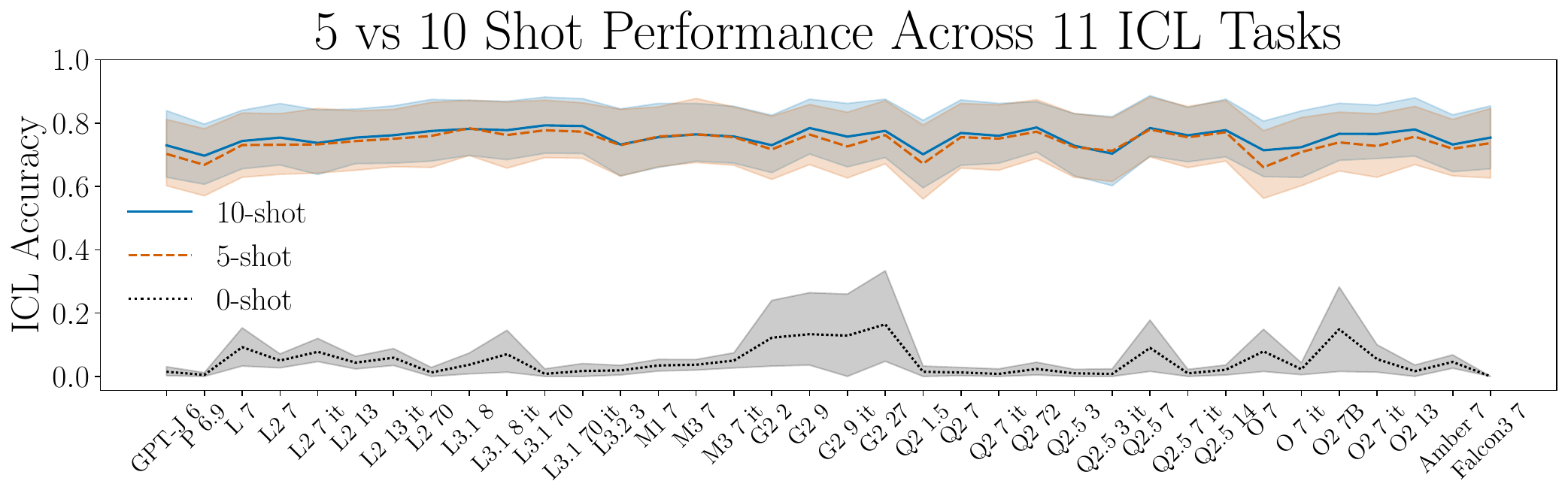}
  \caption{10-shot and 5-shot performance across all models are relatively consistent, with 5-shot under performing by a small margin in some cases. Continue reading in \Aref{ap:ap_evaluation}}
  \label{fig:five_ten_shot}
\end{figure*}

As discussed in \Sref{sec:ap_setup}, we provide the full formulation for calculating peak and average performance recovery in FV and TV.
\begin{align}
    \mathrm{peak} &= \max_{\mathcal{A}_n, \lambda, \ell}(\frac{1}{|P_t|}\sum_{p^t_i \in P_t}\frac{f_{v_t}(p_i^{t,0})=y_i}{f(p_i^{t,5})=y_i}) \\
    \mathrm{avg} &= \max_{\mathcal{A}_n, \lambda}(\frac{1}{L}\sum_{\ell \in L}\frac{1}{|P_t|}\sum_{p^t_i \in P_t}\frac{f_{v_t}(p_i^{t,0})=y_i}{f(p_i^{t,5})=y_i})
\end{align}
Expanding on K shot evaluation in \Sref{sec:ap_setup}, we compare zero-shot, 5-shot and 10-shot performance for all models in \autoref{fig:five_ten_shot}. We find that on average, 10-shot performance slightly beats out 5-shot. By creating FVs with 10-shot examples, and evaluating the recovery according to 5-shot performance, we slightly overestimate the effectiveness of FVs and underestimate its brittleness. That is, to say our findings hold well for both setups.

\subsection{Additional Detail on Activation Patching Results}
\label{ap:ap_results}
\begin{figure*}[t]
  \includegraphics[width=\textwidth]{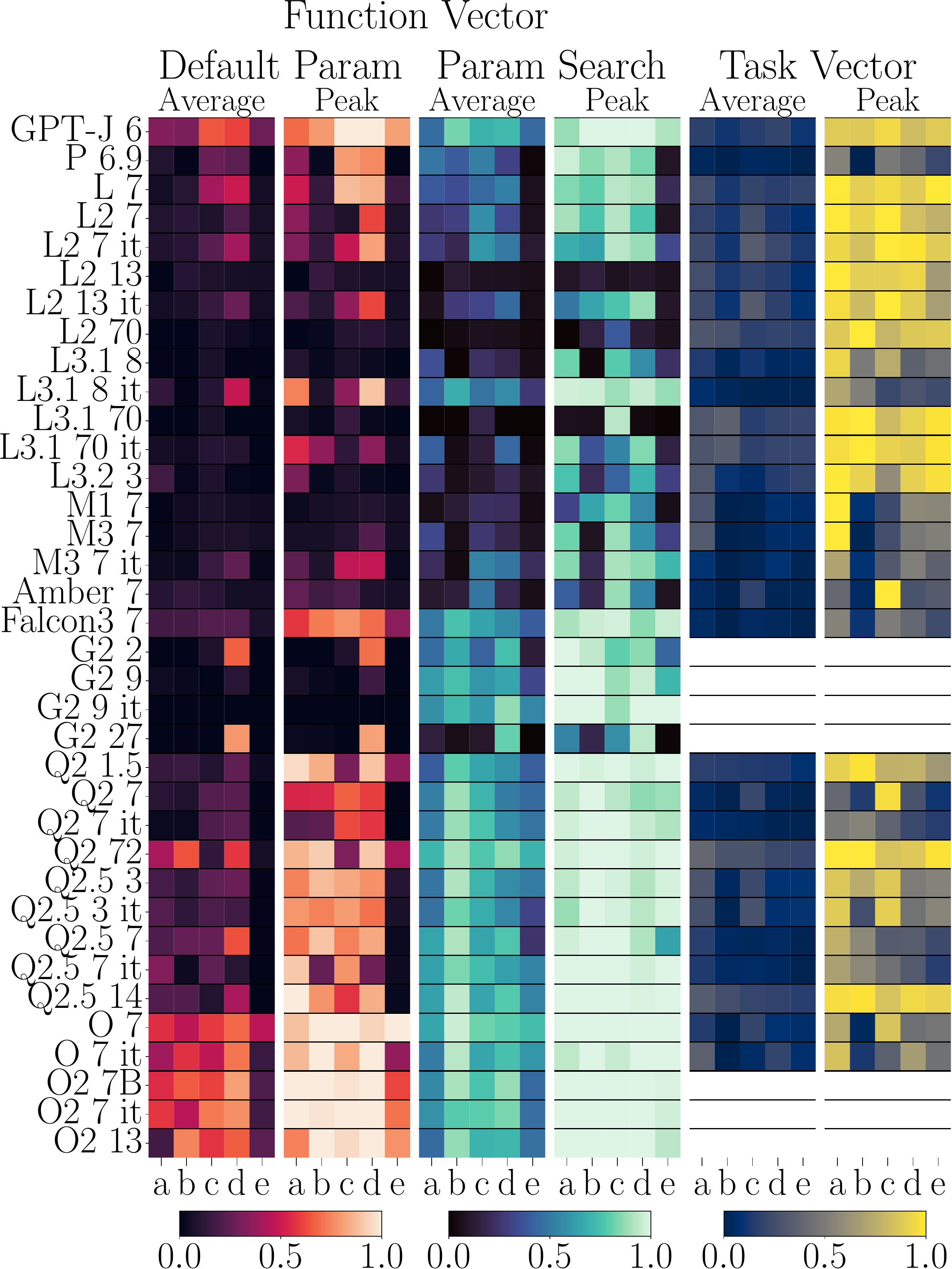}
  \caption{Heatmap showing performance recovery with different activation patching methods across tasks and models. This plot includes OLMo2 and Gemma2, which are omitted from the similar plot in the main paper. There is large variance in performance across tasks, models, and tools. Tasks: a) Antonym, b) Present-Past, c) Country-Capital, d) [lang] to eng, and e) eng to [lang]. Continue reading in \Aref{ap:ap_results}}
  \label{fig:6_plot_all_models}
\end{figure*}

For a comprehensive display of figures from models not displayed in this paper, please follow \href{https://github.com/patqdasilva/steering-off-course}{https://github.com/patqdasilva/steering-off-course}
\paragraph{Additional results with OLMo2 and Gemma2 using FVs}
Corresponding to the results from \Sref{sec:ap_setup}, we show additional results for OLMo2 and Gemma2 in \autoref{fig:6_plot_all_models}. OLMo2 shows decent FV performance similar to OLMo, whereas Gemma2 has varying performance which is especially poor in the 27B model.

\begin{figure}[t]
  \includegraphics[width=\columnwidth]{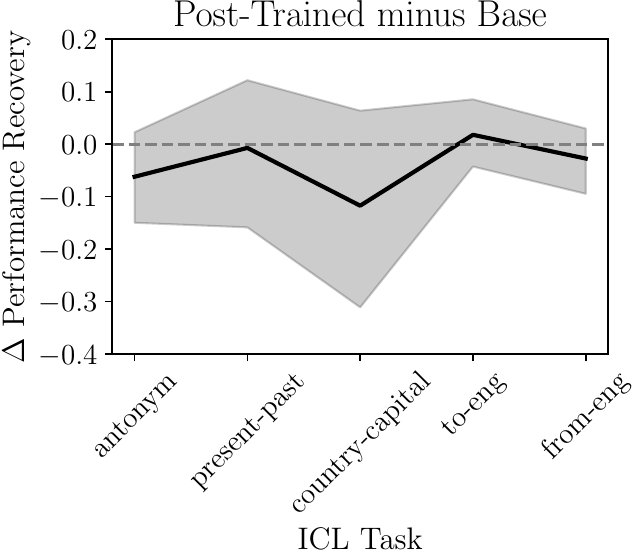}
  \caption{Task vectors performance recovery of post-trained models against their base counterparts. In most cases, post-trained models perform worse or the same when patched with a task vector. Continue reading in \Sref{sec:ap_results}.}
  \label{fig:tv_instr}
\end{figure}

\paragraph{Task Vectors: Post Training}
As discussed in \Sref{sec:ap_results}, we find that unlike FVs, instruction tuned models using TVs are not more steerable than their base versions. We display this result across tasks in \autoref{fig:tv_instr}.

\begin{figure}[ht]
  \includegraphics[width=\columnwidth]{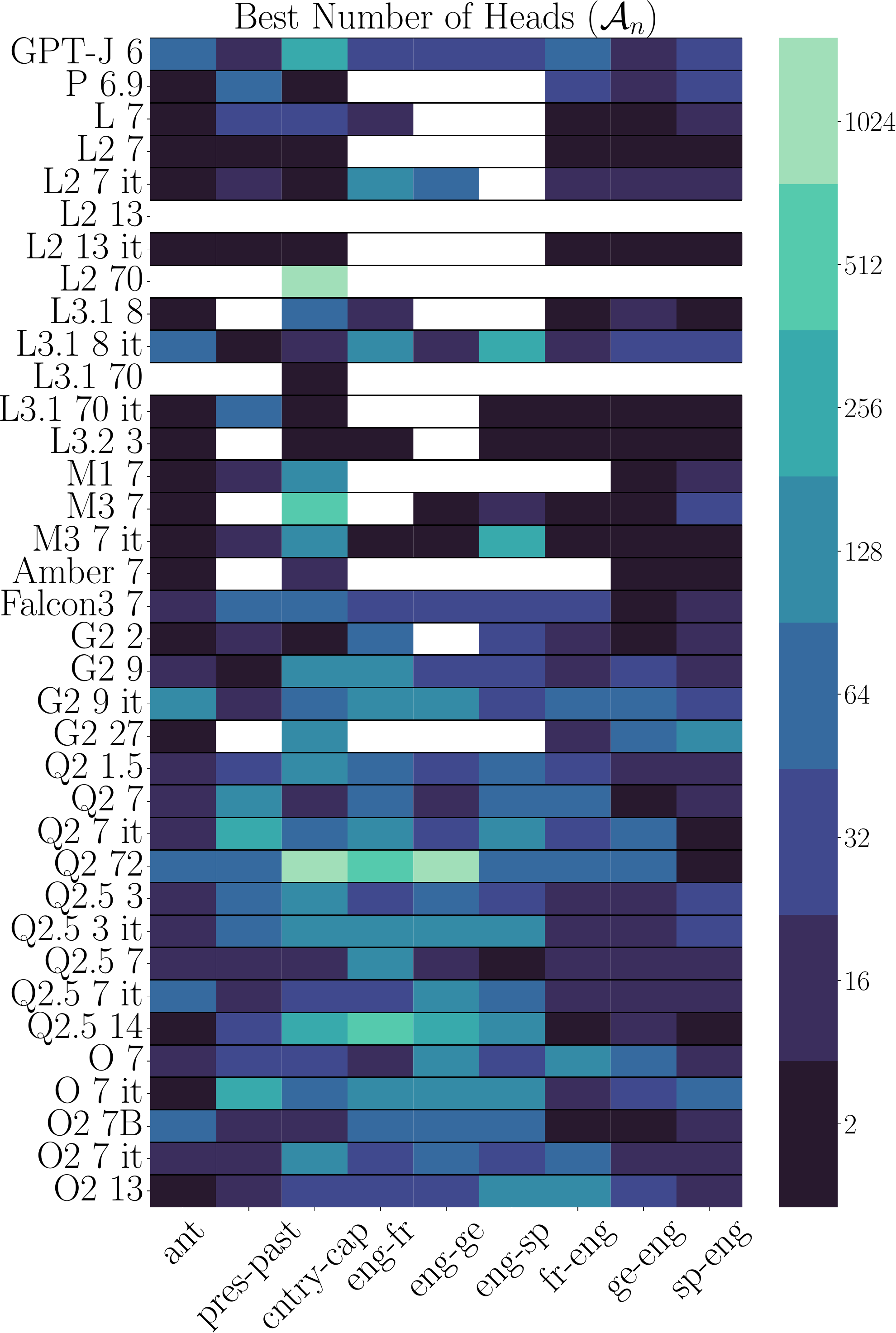}
  \caption{Best number of heads $\mathcal{A}_n$ considering all other hyperparameters for function vectors. Data is masked for models with zero-shot, function vector patched accuracy of at least 10\% on the task. Continue reading in \Sref{sec:ap_results}.}
  \label{fig:best_heads}
\end{figure}

\begin{figure}[ht]
  \includegraphics[width=\columnwidth]{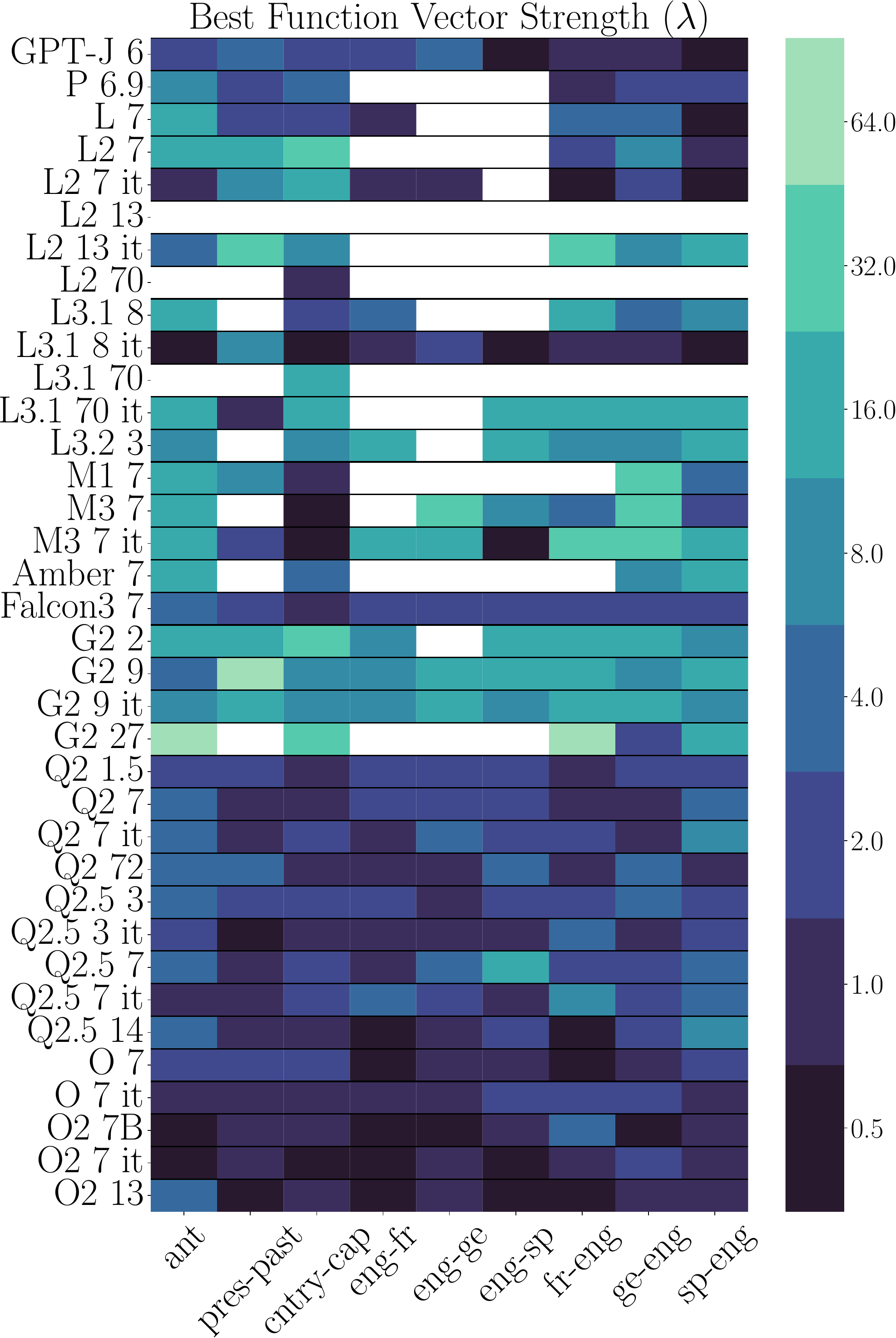}
  \caption{Best function vector strength $\lambda$ considering all other hyperparameters for function vectors. Data is masked for models with zero-shot, function vector patched accuracy of at least 10\% on the task. Continue reading in \Sref{sec:ap_results}.}
  \label{fig:best_lambda}
\end{figure}

\begin{figure}[t]
  \includegraphics[width=\columnwidth]{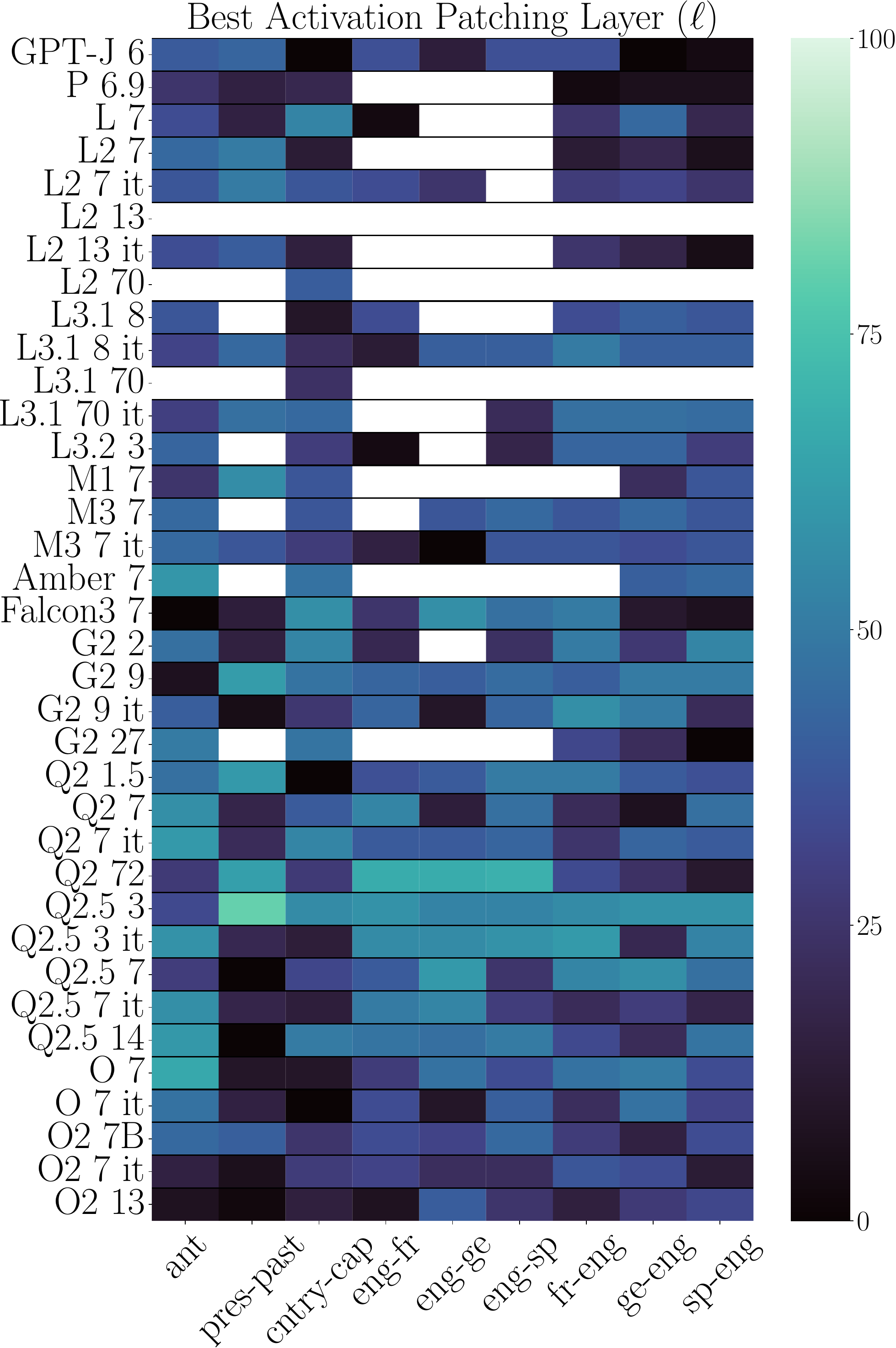}
  \caption{Best layer $\ell$ considering all other hyperparameters for function vectors expressed as a percent of all layers. Data is masked for models with zero-shot, function vector patched accuracy of at least 10\% on the task. Continue reading in \Sref{sec:ap_results}.}
  \label{fig:best_layer}
\end{figure}

\paragraph{Best Hyperparameters}
\label{ap:best_param}
As discussed in \Sref{sec:ap_results}, we show the hyperparameters with best performance across all models and tasks in \autoref{fig:best_heads} (number of heads $\mathcal{A}_n$), \autoref{fig:best_lambda} (strength of the function vector $\lambda$), and \autoref{fig:best_layer} (layer $\ell$).

\begin{figure*}[ht]
    \centering
  \includegraphics[width=0.6\textwidth]{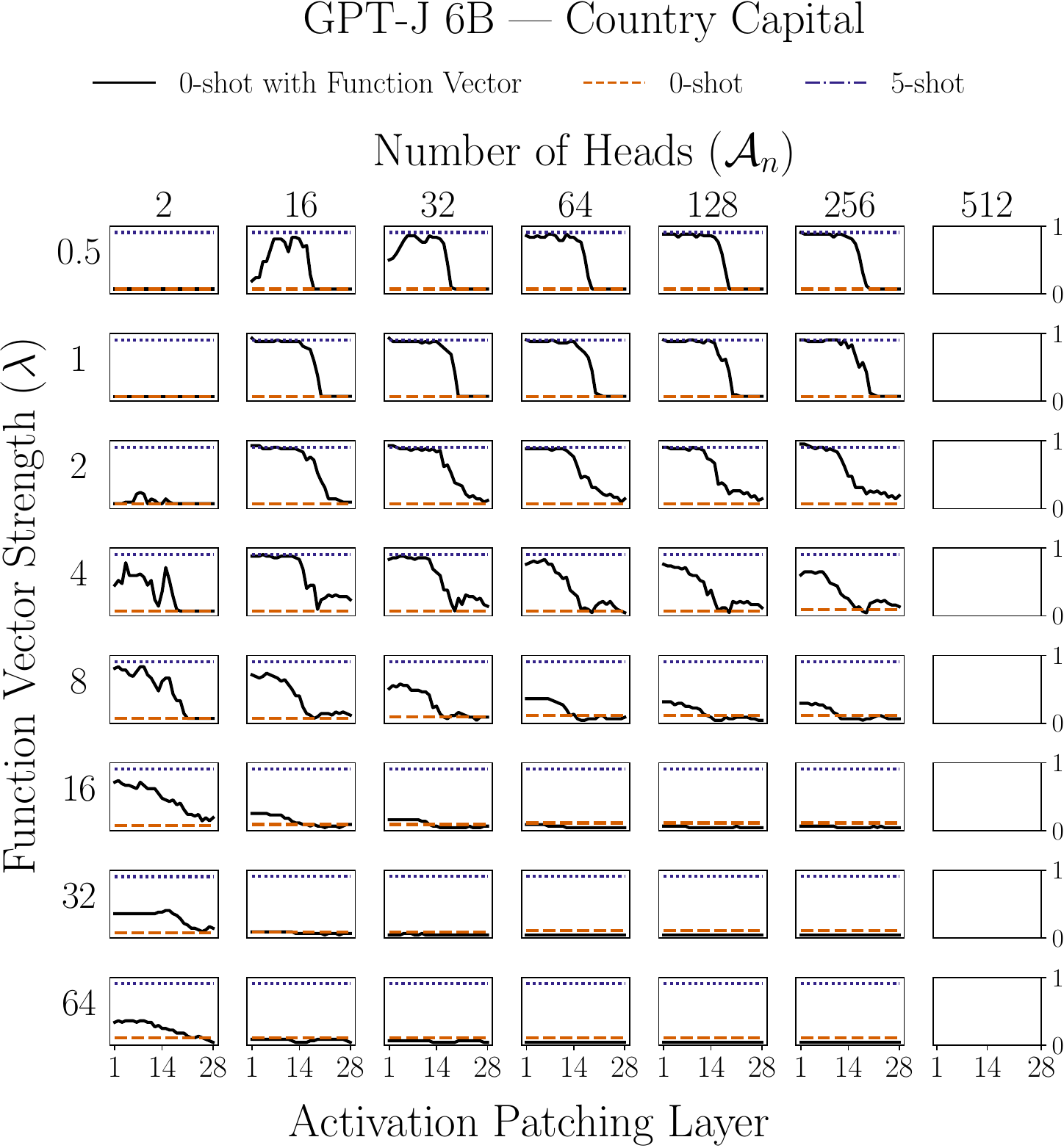}
  \caption{Full parameter search for GPT-J 6B on the country-capital task. Continue reading in \Sref{sec:ap_results}.}
  \label{fig:fv_gptj}
\end{figure*}

\paragraph{Function Vectors: Full Hyperparameter Search}
We show a contrast to the non-localized Mistral-v0.3 7B (Country-Capital task) from \autoref{fig:fv_param_search}. GPT-J 6B shown in \fref{fig:fv_gptj}, which was studied extensively by \citet{todd2024function}, does show evidence for localization in the Country Capital task. It has 448 heads while nearly recovering its maximum performance with the FV built with only 2 (0.44\%) heads.
These two models are just one of several interactions of behavior differences between models that are difficult to parse. We note strong robustness across all hyperparameters. For every model and task tested, please visit \href{https://github.com/patqdasilva/steering-off-course}{https://github.com/patqdasilva/steering-off-course}

\begin{figure*}[!t]
  \includegraphics[width=\textwidth]{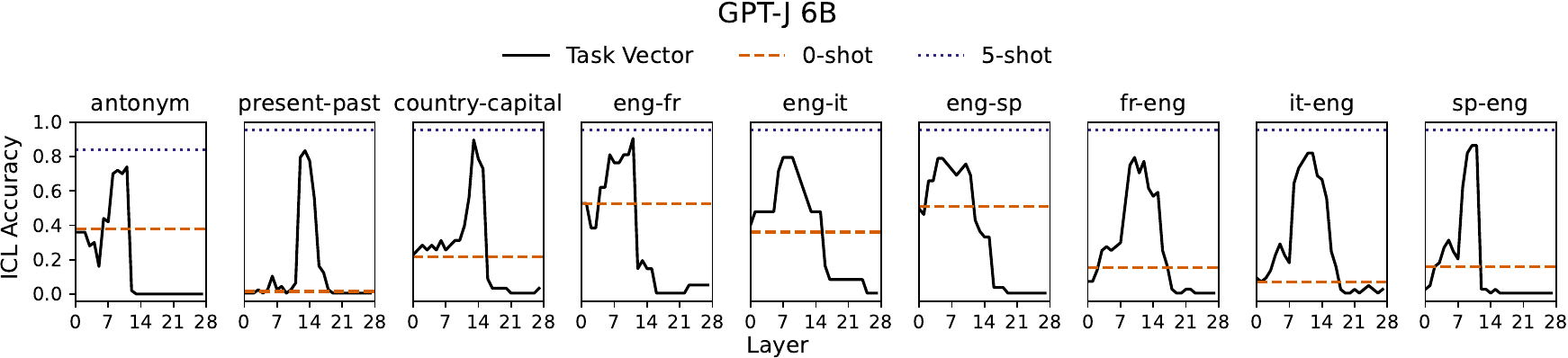}
  \caption{GPT-J 6B accuracy with task vector patched into layer $\ell$ in a zero-shot prompt. Consistent with \citet{hendel2023incontext}, this model has decent performance recovery across all tasks. Continue reading in \Aref{ap:tv_param_search}.}
  \label{fig:tv_gptj}
\end{figure*}

\begin{figure*}[!t]
  \includegraphics[width=\textwidth]{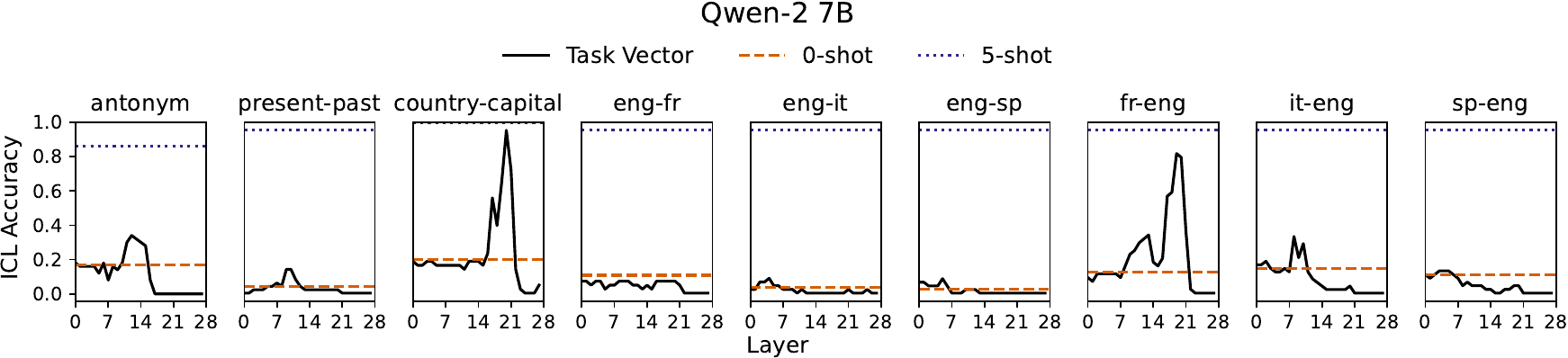}
  \caption{Qwen2 7B ICL accuracy with task vector patched into layer $\ell$ in a zero-shot prompt. Unlike GPT-J, this model fails to recover performance across many of its tasks. Continue reading in \Aref{ap:tv_param_search}.}
  \label{fig:tv_qwen2}
\end{figure*}

\paragraph{Task Vectors: Full Hyperparameter Search}
\label{ap:tv_param_search}
We provide representative examples of full parameter search as dicussed in \Sref{sec:ap_results}. We compare a model replicated form prior work in \autoref{fig:tv_gptj} to a model previously unstudied in \autoref{fig:tv_qwen2}. For every model and task tested, please visit \href{https://github.com/patqdasilva/steering-off-course}{https://github.com/patqdasilva/steering-off-course}

\section{Compute Budget and Hardware}
We use V100s with 16 and 32 GB for smaller models, and A100s and H100s with 80GB and 95GB for larger models. We consider small models to be those $<13B$. The bulk of GPU hours come from the hyperparameter search with function vectors. For our small models, creating the FV takes 40 minutes, and search takes 1.5 hours. For our large models, creating the FV takes 4 hours and search takes 2.5 hours. With 27 smaller models and 9 larger models, across 9 tasks in FVs, this comes out to 524 hours on V100s and 526 hours on A100s and H100s. Total emissions are estimated to be 124.72 kgCO$_2$eq \citep{lacoste2019quantifying}.

\begin{sidewaystable*}[!ht]\centering
\scriptsize
\begin{tabular}{lccccccccccccc}\toprule
&\multicolumn{12}{c}{\textbf{Architecture}} \\\cmidrule{2-13}
\textbf{Model Name} &n\_layers &n\_heads &total\_heads &head\_size &hidden\_dim &mlp\_size &layer\_norm (eps) &embed ($\theta$) &attention &hidden\_act &vocab\_size &context\_len \\\midrule
EleutherAI/gpt-j-6b &28 &16 &448 &256 &4,096 &16,384 &LayerNorm (1e-05) &RoPE &MHA &gelu\_new &50,400 &2,048 \\
EleutherAI/pythia-6.9b &32 &32 &1,024 &128 &4,096 &16,384 &LayerNorm (1e-05) &RoPE (10k) &MHA &gelu &50,432 &2,048 \\
huggyllama/llama-7b &32 &32 &1,024 &128 &4,096 &11,008 &LayerNorm (1e-06) &RoPE &MHA &SwiGLU &32,000 &2,048 \\
meta-llama/Llama-2-7b-hf &32 &32 &1,024 &128 &4,096 &11,008 &LayerNorm (1e-05) &RoPE &MHA &SwiGLU &32,000 &4,096 \\
meta-llama/Llama-2-7b-chat-hf &32 &32 &1,024 &128 &4,096 &11,008 &LayerNorm (1e-05) &RoPE &MHA &SwiGLU &32,000 &4,096 \\
meta-llama/Llama-2-13b-hf &40 &40 &1,600 &128 &5,120 &13,824 &LayerNorm (1e-05) &RoPE &MHA &SwiGLU &32,000 &4,096 \\
meta-llama/Llama-2-13b-chat-hf &40 &40 &1,600 &128 &5,120 &13,824 &LayerNorm (1e-05) &RoPE &MHA &SwiGLU &32,000 &4,096 \\
meta-llama/Llama-2-70b &80 &64 &5,120 &128 &8,192 &28,672 &LayerNorm (1e-05) &RoPE &GQA &SwiGLU &32,000 &4,096 \\
meta-llama/Meta-Llama-3-8B &32 &32 &1,024 &128 &4,096 &14,336 &LayerNorm (1e-05) &RoPE (500k) &GQA &SwiGLU &128,256 &8,192 \\
meta-llama/Meta-Llama-3-70B &80 &64 &5,120 &128 &8,192 &28,672 &LayerNorm (1e-05) &RoPE (500k) &GQA &SwiGLU &128,256 &8,192 \\
meta-llama/Llama-3.1-8B &32 &32 &1,024 &128 &4,096 &14,336 &LayerNorm (1e-05) &RoPE (500k) &GQA &SwiGLU &128,256 &131,072 \\
meta-llama/Llama-3.1-8B-Instruct &32 &32 &1,024 &128 &4,096 &14,336 &LayerNorm (1e-05) &RoPE (500k) &GQA &SwiGLU &128,256 &131,072 \\
meta-llama/Llama-3.1-70B &80 &64 &5,120 &128 &8,192 &28,672 &LayerNorm (1e-05) &RoPE (500k) &GQA &SwiGLU &128,256 &131,072 \\
meta-llama/Llama-3.1-70B-Instruct &80 &64 &5,120 &128 &8,192 &28,672 &LayerNorm (1e-05) &RoPE (500k) &GQA &SwiGLU &128,256 &131,072 \\
meta-llama/Llama-3.2-3B &28 &24 &672 &128 &3,072 &8,192 &LayerNorm (1e-05) &RoPE (500k) &GQA &SwiGLU &128,256 &131,072 \\
mistralai/Mistral-7B-v0.1 &32 &32 &1,024 &128 &4,096 &14,336 &LayerNorm (1e-05) &RoPE &GQA &SwiGLU &32,000 &8,192 \\
mistralai/Mistral-7B-v0.3 &32 &32 &1,024 &128 &4,096 &14,336 &LayerNorm (1e-05) &RoPE &GQA &SwiGLU &32,768 &8,192 \\
LLM360/Amber &32 &32 &1,024 &128 &4,096 &11,008 &LayerNorm (1e-06) &RoPE &MHA &SwiGLU &32,000 &2,048 \\
tiiuae/Falcon3-7B-Base &28 &12 &336 &256 &3,072 &23,040 &LayerNorm (1e-06) &RoPE (1,000k) &GQA &SwiGLU &131,072 &32,768 \\
Qwen/Qwen2-1.5B &28 &12 &336 &128 &1,536 &8,960 &LayerNorm (1e-06) &RoPE (1,000k) &GQA &SwiGLU &151,646 &131,072 \\
Qwen/Qwen2-7B &28 &28 &784 &128 &3,584 &18,944 &LayerNorm (1e-06) &RoPE (1,000k) &GQA &SwiGLU &151,646 &131,072 \\
Qwen/Qwen2-7B-Instruct &28 &28 &784 &128 &3,584 &29,568 &LayerNorm (1e-06) &RoPE (1,000k) &GQA &SwiGLU &151,646 &131,072 \\
Qwen/Qwen2-72B &80 &64 &5,120 &128 &8,192 &29,568 &LayerNorm (1e-06) &RoPE (1,000k) &GQA &SwiGLU &151,646 &131,072 \\
Qwen/Qwen2.5-3B &36 &16 &576 &128 &2,048 &11,008 &LayerNorm (1e-06) &RoPE (1,000k) &GQA &SwiGLU &151,936 &32,768 \\
Qwen/Qwen2.5-3B-Instruct &36 &16 &576 &128 &2,048 &11,008 &LayerNorm (1e-06) &RoPE (1,000k) &GQA &SwiGLU &151,936 &131,072 \\
Qwen/Qwen2.5-7B &28 &28 &784 &128 &3,584 &18,944 &LayerNorm (1e-06) &RoPE (1,000k) &GQA &SwiGLU &152,064 &131,072 \\
Qwen/Qwen2.5-7B-Instruct &28 &28 &784 &128 &3,584 &18,944 &LayerNorm (1e-06) &RoPE (1,000k) &GQA &SwiGLU &152,064 &131,072 \\
Qwen/Qwen2.5-14B &48 &40 &1,920 &128 &5,120 &13,824 &LayerNorm (1e-05) &RoPE (1,000k) &GQA &SwiGLU &152,064 &131,072 \\
google/gemma-2-2b &26 &8 &208 &256 &2,304 &18,432 &LayerNorm (1e-06) &RoPE (10k) &GQA &GeGLU &256,128 &8,192 \\
google/gemma-2-9b &42 &16 &672 &256 &3,584 &28,672 &LayerNorm (1e-06) &RoPE (10k) &GQA &GeGLU &256,128 &8,192 \\
google/gemma-2-9b-it &42 &16 &672 &256 &3,584 &28,672 &LayerNorm (1e-06) &RoPE (10k) &GQA &GeGLU &256,128 &8,192 \\
google/gemma-2-27b &46 &32 &1,472 &128 &4,608 &73,728 &LayerNorm (1e-06) &RoPE (10k) &GQA &GeGLU &256,128 &8,192 \\
allenai/OLMo-7B-0724-hf &32 &32 &1,024 &128 &4,096 &11,008 &non-parametric &RoPE (10k) &MHA &SwiGLU &50,304 &4,096 \\
allenai/OLMo-7B-0724-Instruct-hf &32 &32 &1,024 &128 &4,096 &11,008 &non-parametric &RoPE (10k) &MHA &SwiGLU &50,304 &4,096 \\
allenai/OLMo-2-1124-7B &32 &32 &1,024 &128 &4,096 &11,008 &LayerNorm (1e-06) &RoPE (500k) &MHA &SwiGLU &100,352 &4,096 \\
allenai/OLMo-2-1124-7B-Instruct &32 &32 &1,024 &128 &4,096 &11,008 &LayerNorm (1e-06) &RoPE (500k) &MHA &SwiGLU &100,352 &4,096 \\
allenai/OLMo-2-1124-13B &40 &40 &1,600 &128 &5,120 &13,824 &LayerNorm (1e-06) &RoPE (500k) &MHA &SwiGLU &100,352 &4,096 \\
\bottomrule
\end{tabular}
\caption{Architecture differences among the models tested. Continue reading in \Sref{sec:discussion}.}\label{tab:model_arch}
\end{sidewaystable*}

\begin{sidewaystable*}[!ht]\centering
\scriptsize
\begin{tabular}{lrrrrr}\toprule
\textbf{} &\multicolumn{4}{c}{\textbf{Data, Training, and Optimization}} \\\cmidrule{2-5}
\textbf{Model Name} &n\_tokens &pt\_learn\_rate &staged\_pretrain &post\_trained \\\midrule
EleutherAI/gpt-j-6b &402B &- &no &no \\
EleutherAI/pythia-6.9b &300B &1.2 x 10-4 &no &no \\
huggyllama/llama-7b &1T &3 × 10-4 &no &no \\
meta-llama/Llama-2-7b-hf &2T &3 × 10-4 &no &no \\
meta-llama/Llama-2-7b-chat-hf &2T &3 × 10-4 &no &yes \\
meta-llama/Llama-2-13b-hf &2T &3 × 10-4 &no &no \\
meta-llama/Llama-2-13b-chat-hf &2T &3 × 10-4 &no &yes \\
meta-llama/Llama-2-70b &2T &1.5 × 10-4 &no &no \\
meta-llama/Meta-Llama-3-8B &15T+ &3 × 10-4 &no &no \\
meta-llama/Meta-Llama-3-70B &15T+ &1.5 × 10-4 &no &no \\
meta-llama/Llama-3.1-8B &15T+ &3 × 10-4 &yes &no \\
meta-llama/Llama-3.1-8B-Instruct &15T+ &3 × 10-4 &yes &yes \\
meta-llama/Llama-3.1-70B &15T+ &1.5 × 10-4 &yes &no \\
meta-llama/Llama-3.1-70B-Instruct &15T+ &1.5 × 10-4 &yes &yes \\
meta-llama/Llama-3.2-3B &15T+ &- &yes &no \\
mistralai/Mistral-7B-v0.1 &- &- &no &no \\
mistralai/Mistral-7B-v0.3 &- &- &no &no \\
LLM360/Amber &1.3T &3 × 10-4 &no &no \\
tiiuae/Falcon3-7B-Base &14T &- &- &no \\
Qwen/Qwen2-1.5B &7T+ &- &yes &no \\
Qwen/Qwen2-7B &7T+ &- &yes &no \\
Qwen/Qwen2-7B-Instruct &7T+ &- &yes &yes \\
Qwen/Qwen2-72B &7T+ &- &yes &no \\
Qwen/Qwen2.5-3B &18T &- &yes &no \\
Qwen/Qwen2.5-3B-Instruct &18T &- &yes &yes \\
Qwen/Qwen2.5-7B &18T &- &yes &no \\
Qwen/Qwen2.5-7B-Instruct &18T &- &yes &yes \\
Qwen/Qwen2.5-14B &18T &- &yes &no \\
google/gemma-2-2b &2T &- &no &no \\
google/gemma-2-9b &8T &- &no &no \\
google/gemma-2-9b-it &8 &- &no &yes \\
google/gemma-2-27b &8T &- &no &no \\
allenai/OLMo-7B-0724-hf &2.75T & 3 x 10-4 &yes &no \\
allenai/OLMo-7B-0724-Instruct-hf &2.75T &3 × 10-4 &yes &yes \\
allenai/OLMo-2-1124-7B &5T &3 × 10-4 &yes &no \\
allenai/OLMo-2-1124-7B-Instruct &5T &3 × 10-4 &yes &yes \\
allenai/OLMo-2-1124-13B &5T &9 × 10-4 &yes &yes \\
\bottomrule
\end{tabular}
\caption{Data, training, and optimization differences among the models tested. "-" represents information which was not found, not published, or unclear. Continue reading in \Sref{sec:discussion}}\label{tab:model_dto}    
\end{sidewaystable*}

\end{document}

%% file: macros.tex
\newcommand{\missingcitation}[1]{\textcolor{red}{[CITE]}}
\newcommand{\missingnumber}[1]{\textcolor{red}{[NUMBER]}}

\newcommand{\Sref}[1]{\S\ref{#1}}
\newcommand{\fref}[1]{Figure~\ref{#1}}

\newcommand{\Aref}[1]{Appendix~\ref{#1}}

\usepackage{xspace}

\usepackage{enumitem}

\definecolor{codegreen}{rgb}{0,0.6,0}
\definecolor{codegray}{rgb}{0.5,0.5,0.5}
\definecolor{codepurple}{rgb}{0.58,0,0.82}
\definecolor{backcolour}{rgb}{0.95,0.95,0.92}

\newcommand{\applygradient}[1]{%
  \pgfmathparse{#1}%
  \let\val=\pgfmathresult 
  \pgfmathparse{int(\val)} 
  \let\intval=\pgfmathresult 
  \ifnum\intval<50
    \pgfmathsetmacro{\r}{255}
    \pgfmathsetmacro{\g}{128 * \val / 50}
  \else
    \pgfmathsetmacro{\r}{255 * (100 - \val) / 50}
    \pgfmathsetmacro{\g}{255}
  \fi
  \pgfmathtruncatemacro{\rint}{\r}%
  \pgfmathtruncatemacro{\gint}{\g}%
  \xdef\colorname{\noexpand\cellcolor[RGB]{\rint,\gint,0}}
  \colorname
  \val
}